\documentclass[journal]{IEEEtran}
\usepackage{graphicx}%figure package
\usepackage{float}
\usepackage{subfigure}
\usepackage{algorithm,algorithmic}
\usepackage{dsfont}
\usepackage{amssymb}
\usepackage{amsmath}
\usepackage{url}
\usepackage{multirow}
\usepackage{booktabs}%
\usepackage{threeparttable}

\usepackage{cite}
\usepackage{color}
\ifCLASSINFOpdf
\else
\fi
\begin{document}
\title{Domain-invariant Prototypes for Semantic Segmentation}
%in part by the Project of Science Fund for Distinguished Young Scholars of Hunan Province under Grant 2021JJ10024, in part by the Key Research and Development Project of Science and Technology Plan of Hunan Province under Grant 2022GK2014, in part by China Postdoctoral Science Foundation under Grant 2021M701156, in part by the Project of Talent innovation and Sharing Alliance of Quanzhou City under Grant 2021C062L.

\author{Zhengeng~Yang,
        Hongshan~Yu,
        Wei~Sun,
        Li-Cheng,
        Ajmal~Mian% <-this % stops a space
\thanks{This work was supported in part by the National Natural Science Foundation of China under Grant 62103137, U2013203, 61973106, U1913202,  Professor Ajmal Mian is the recipient of an Australian Research Council Future Fellowship Award (project number FT210100268) funded by the Australian Government.}
% <-this % stops a space

\thanks{Zhengeng Yang, Hongshan Yu,  and Wei Sun are with the National Engineering Laboratory for Robot Visual Perception and Control Technology, College of Electrical and Information Engineering, Hunan University, Lushan South Rd., Yuelu Dist., 410082, Changsha, China. Zhengeng Yang and Hongshan Yu contributed equally to this work.  H. Yu is the corresponding author ( e-mail: yuhongshancn@hotmail.com).}
\thanks{Li Cheng is with the Department of Electrical and Computer Engineering, University of Alberta, Edmonton, AB, Canada.}
\thanks{Ajmal~Mian is with the Department of Computer Science, The University
of Western Australia, Perth, WA 6009, Australia}}

%and in part by the Project of Talent Innovation and Sharing Alliance of Quanzhou City under Grant 2021C062L



\markboth{Journal of \LaTeX\ Class Files,~Vol.~14, No.~8, November~2018}%
{Shell \MakeLowercase{\textit{et al.}}: Bare Demo of IEEEtran.cls for IEEE Journals}



\maketitle

\begin{abstract}
Deep Learning has greatly advanced the performance of semantic segmentation, however, its success relies on the availability of large amounts of annotated data for training. Hence, many efforts have been devoted to domain adaptive semantic segmentation that focuses on transferring semantic knowledge from a labeled source domain to an unlabeled target domain. Existing self-training methods typically require multiple rounds of training, while another popular framework based on adversarial training is known to be sensitive to hyper-parameters. In this paper, we present an easy-to-train framework that learns domain-invariant prototypes for domain adaptive semantic segmentation. In particular, we show that domain adaptation shares a common character with few-shot learning in that both aim to recognize some types of unseen data with knowledge learned from large amounts of seen data. Thus, we propose a unified framework for domain adaptation and few-shot learning. The core idea is to use the class prototypes extracted from few-shot annotated target images to classify pixels of both source images and target images. Our method involves only one-stage training and does not need to be trained on large-scale un-annotated target images. Moreover, our method can be extended to variants of both domain adaptation and few-shot learning. Experiments on adapting GTA5-to-Cityscapes and SYNTHIA-to-Cityscapes show that our method achieves competitive performance to state-of-the-art.

\end{abstract}




\section{Introduction}
\label{sec_intro}
\IEEEPARstart{S}{emantic} segmentation is the task of assigning each pixel a semantic label, such as car, bicycle, tree, among others. Recent deep learning methods (e.g.~~\cite{yang2020small,yang2020ndnet}) have significantly advanced the state-of-the-art performance in semantic segmentation. However, their performance heavily relies on the availability of large-scale labelled training datasets. Unfortunately, for semantic segmentation, this entails pixel-level annotations that is often an extremely time consuming and tedious process. For example, Saleh et al.~\cite{saleh2018effective} mention that it takes on average 90 minutes to manually annotate a single $1024\times2048$ image for semantic segmentation.

Numerous research attempts~\cite{cordts2016cityscapes,ros2016synthia} have been made to avoid the manual annotation by leveraging computer graphics techniques to automatically generate synthetic images along with their corresponding pixel-wise labels. However, models directly trained on synthetic images often do not perform well on real data, which is largely attributed to the issue of domain shift. Therefore,  domain adaptive
semantic segmentation that focuses on transferring semantic knowledge from a labeled source domain to an unlabeled target domain has attracted significant research attention over the past decade~\cite{li2020content,zheng2021rectifying,zhang2021prototypical,wang2021uncertainty,araslanov2021self}. Although current self-training based methods and adversarial training based methods methods have achieved great success in domain adaptive semantic segmentation,  these two popular approaches usually involve a cumbersome training process. In particular, both need to pre-train an initial segmentation model on the source domain. Additionally, self-training based methods generally perform self-training on the entire target domain for multiple rounds, while GAN-based methods are known to be difficult to train due to their sensitivity to hyper-parameters.

\begin{figure}[!t]
\centering
\includegraphics[width=3.3in]{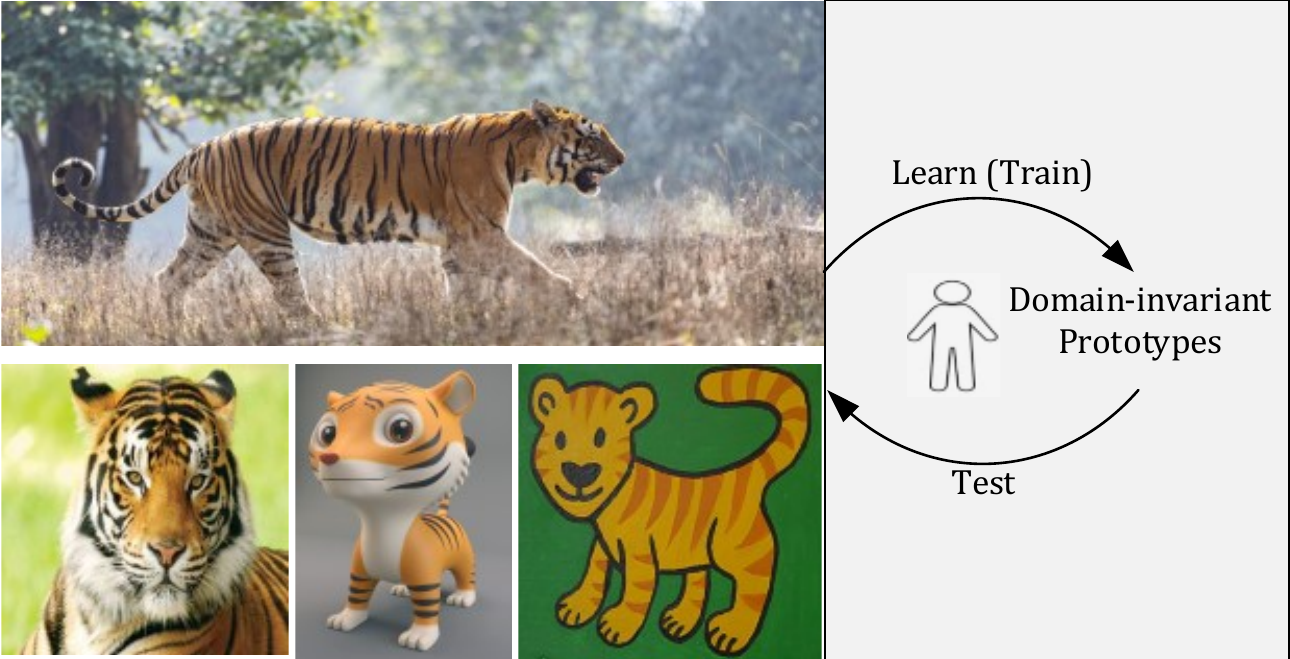}
\caption{The main hypothesis that motivates our method. %We hypothesize the magical-like domain adaptation ability of human originated from that human can learn domain-invariant class prototypes, and the object recognition of human is more likely a ``similarity comparing" process with respect to these prototypes.
The impressive domain adaptation ability of humans (we can still recognize the second row images as tiger) originates from the fact that we learn domain-invariant class prototypes, and then perform object classification through a ``similarity comparison" process with respect to the prototypes. }
\label{fig_tiger}
\end{figure}

Besides adaptation from virtual domain, another direction of alleviating the reliance on large-scale manual annotations is to use the few-shot learning, which aims to learn to recognize some query classes with only a few (e.g., 1,5) annotated support samples of them. Towards this goal, the few-shot learning typically learns the model over a set of base classes that has already been associated with a large number of annotated samples, while the challenging few-shot scenario is only applied to unseen classes. Since the few-shot learning provides a solution to extend the learned model to unseen classes, it has received great attention in recent several years, especially the prototype-based method~\cite{siam2019amp,li2021adaptive,liu2020crnet}. The core idea of prototype based few-shot learning is to treat the average features of the few-shot support samples as the prototypes of unseen classes. Then, the classification of the query samples can be performed by measuring the feature distances to these prototypes and choosing the nearest class prototype as the result. 

%It can be observed that both the domain adaptation and few-shot learning aim to recognize some unseen data with knowledge learned from seen data. In particular, domain adaptation aims to recognize unseen data distributions while few-shot learning aims to recognize unseen classes. In other words, if we view all the classes presented in unseen distribution in the target domain as unseen classes to the source domain, and if a few target annotations are available at the same time, the domain adaptive semantic segmentation can be also transformed into a few-shot semantic segmentation problem. 

\begin{figure*}[!ht]
\centering
\includegraphics[width=\textwidth]{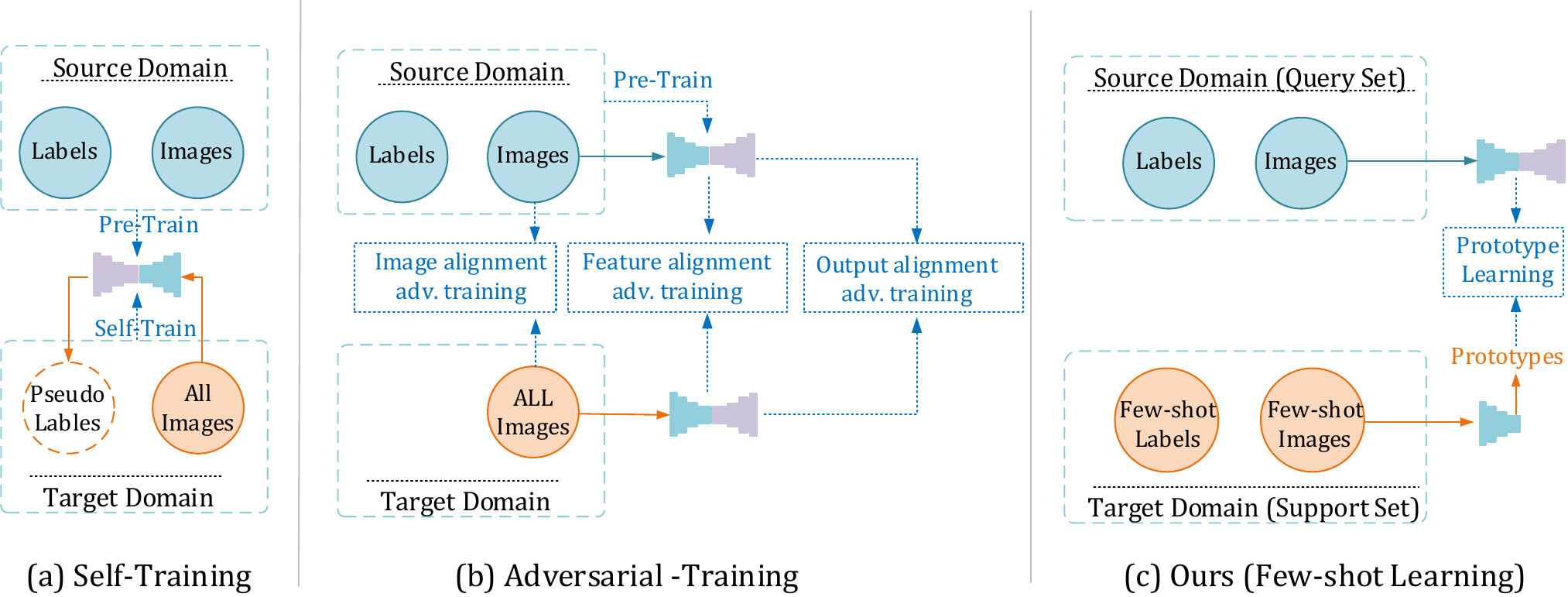}
\caption{The proposed framework compared to two popular frameworks. Self-training and adversarial-training methods generally involve cumbersome training processes. The former usually needs to perform self-training on the entire target dataset for multiple rounds, while the latter needs to optimize multiple adversarial losses which are sensitive to hyper parameters. In contrast, our proposed method learns domain-invariant prototypes with the support of few-shot target annotations. Our method involves only one-stage training and need not to access large-scale un-annotated target images.}
\label{fig_difference}
\end{figure*}

Inspired by the success of prototype-based class representation in few-shot learning, we hypothesize that the impressive domain adaptation ability of humans (Fig. \ref{fig_tiger}) originates from the fact that we learn domain-invariant class prototypes, and then perform object classification through a ``similarity comparison" process with respect to the prototypes. Based on this, we propose to formulate the domain adaptation as a domain-invariant prototype learning problem. Toward this goal, we first extract class prototypes from a few target images with annotations, and then use them to segment the source images. The prototype learning thus can be supervised by the large amount of source annotations. After that, the segmentation on arbitrary target images is implemented by computing pixel-wise similarity with these learned class prototypes. As shown in Fig. \ref{fig_difference}, compared with using self-training and adversarial training, our few-shot domain adaptation involves only one stage training and need not to be pre-trained on the source domain. Moreover, our method also need not to access large scale un-annotated target images.

Since our few-shot domain adaption is developed from current popular prototype-based few-shot segmentation framework, it is noteworthy that existing few-shot segmentation methods usually consider a binary segmentation task on a specific class during training.  The reasons for this consideration will be detailed in Section \ref{sec_pre_fewshot}. Although this training strategy works well for low-resolution images containing a few salient objects (e.g., images in PASCAL dataset),  it is not suitable for high-resolution images containing arbitrary number of classes for two reasons. Firstly, there are many object classes (e.g., traffic light) that occupy only a small fraction of the entire image. Binary segmentation on these classes means all other objects will be treated as background, which is harmful for the feature representation learning. Secondly, it is difficult to ensure those small object classes are still contained in the training image after the common random image cropping data augmentation is applied to images of high-resolution. In other words, the classes contained in the training image cropped from high-resolution images are uncertain. Thus, we cannot perform a binary segmentation on a specific class. Based on the observation that the prototypes used for class recognition are extracted from support images, we present a support image depended training strategy in which a class will be recognized if and only if it appears in current support images. Thus, the classes to be recognized in each training step are essentially adapted to the randomly sampled training images, which makes our few-shot learning method can be generalized to most  scenes without constraints on image resolution and saliency of objects. 

%Moreover, the number of few-shot samples for these classes to be recognized are also determined by current support images and not required to keep the same as the one (e.g., 1) used for test stage. For example, if there are $n$ annotated samples of class $A$ in current support images, all the $n$ samples will be used to compute the prototypes of class $A$. Since the support images are randomly sampled from the training set and , it 

%the classes as well as their annotated samples used in each training step are fully depended on current support images. In particular,
%we recognize all the classes contained in current support images during training. 

 To summarize, our contributions are three folds as follows.

First, we present a novel domain adaptation framework that formulates domain adaptation as a domain-invariant prototype learning problem. Compared with current methods based on self-training and generative adversarial networks (GAN)~\cite{goodfellow2014generative}, our method is much easier to implement since it did not require to either perform multiple rounds training  or optimize multiple adversarial losses.  

%compared with current binary segmentation based training usually  requires objects occupy a certain proportion of the training image,
Second, we extend the current few-shot segmentation problem to a more generalized form. Instead of segmenting only ``things" (objects) from the background like most few-shot segmentation methods do, the proposed method considers segmenting  ``things" and ``stuffs" simultaneously. Moreover, we present a support image adaptive training strategy for few-shot learning without constraints on image resolution and saliency of objects. To the best of our knowledge, we are the first to consider few-shot segmentation task for high resolution images. 

Third, we developed a unified framework for domain adaptation and few-shot learning, where unified means our method can be applied to both of these two tasks.

%\IEEEpubidadjcol

%The rest of this paper is organized as follow.

\section{Related Work}
\label{sec_relatedwork}

\subsection{Domain Adaptive Semantic Segmentation}
To alleviate the requirement for human-labeled images, a recent trend for semantic segmentation is to use synthetic images where dense annotations are automatically available~\cite{ma2021coarse,zhang2021prototypical,araslanov2021self,fleuret2021uncertainty}. However, models trained on synthetic data when applied directly to real data do not perform well due to the distribution shift between the two domains. This gives rise to the  challenging problem of domain adaptive semantic segmentation. In general, existing domain adaptive semantic segmentation methods can be roughly divided into two categories: self-training based methods and adversarial training based methods.

\noindent \textbf{Self-training based methods:} The core idea of this line of methods is to generate pseudo labels in the target domain, where the pseudo labels are usually obtained by taking the predicted labels with high confidence. Zou et al.~\cite{zou2018unsupervised} were among the first to apply self-training to domain adaptive semantic segmentation. They organized the self-training into a two-stage process (which we name as select-and-train) which is repeated for multiple rounds. Many following works focus on improving the quality of pseudo labels ~\cite{li2020content,zheng2021rectifying,zhang2021prototypical,wang2021uncertainty}. For examples, Li et al.~\cite{li2020content} proposed to use only synthetic images that share similar distribution with the real images for domain adaptation, Zheng et al.~\cite{zheng2021rectifying} developed an uncertainty-based adaptive threshold for pseudo label selection. Since the confidence based pseudo label selection can easily ignore hard samples, another interest of self-training based methods is to prevent the training process being dominated by easy samples~\cite{chen2019domain,zou2019confidence}. Typical ways for solving this problem include designing robust losses~\cite{chen2019domain} and employing multiple classifiers~\cite{zheng2021rectifying}.

%To avoid the cumbersome multiple training rounds, Araslanov et al.~\cite{araslanov2021self} presented a one-round self-training framework by extracting pseudo supervision from semantic consistency between different image transformations.

\noindent \textbf{Adversarial training based methods:} These methods employ adversarial training to align the source domain and target domain either at feature-level, image-level, or both. Adversarial feature alignment generally follows the GAN~\cite{goodfellow2014generative} framework. In particular, by treating the feature encoder as the generator, adversarial feature alignment~\cite{tsai2018learning,hong2018conditional,du2019ssf} tries to fool a discriminator that distinguishes whether the feature is from the source domain or the target domain. Instead of aligning intermediate features, image-level alignment~\cite{wu2018dcan} directly mitigates the domain gap in the image space with image-to-image translation~\cite{zhu2017unpaired}.  More recently, aligning jointly image space, intermediate features, and output space~\cite{chang2019all,yang2021context,yang2020phase} is becoming popular. Although adversarial training provides an effective way to mitigate the domain gap problem, it suffers from  several drawbacks such as being computationally expensive and difficult to train given its sensitivity to hyper-parameters.

\subsection{Few-shot Supervised Domain Adaptation}
Most of the above methods consider an unsupervised domain adaptation problem, i.e., they assume that there are no labeled target images. Inspired by few-shot learning, we relax this setting and assume that there are a few labeled target images available and then propose a few-shot domain adaptation framework. Note that there are some few-shot unsupervised domain adaptation methods~\cite{yue2021prototypical,xu2022few} in which the few-shot scenario is applied over the source domain, i.e., there are only a few labeled source images available and the target domain is still unlabeled. This paper essentially considers a few-shot supervised domain adaption problem that has also been explored in many tasks. For example, Motiian et al.~\cite{motiian2017few} developed a few-shot adversarial domain adaptation method for image classification. For the semantic segmentation task, Zhang et al.\cite{zhang2019few} proposed to use a few target annotations to enhance the semantic feature learning on target domain during domain adversarial learning. Thus, these related methods can be still categorized into adversarial training based methods mentioned above. In contrast, we propose to solve the few-shot domain adaptation by the commonly used prototype-based few-shot learning framework. To the best of our knowledge, we are the first to formulate the domain adaption as a few-shot learning problem. 

%Teshima et al.~\cite{teshima2020few} provided a transfer mechanism for regression model adaptation falling into the few-shot supervised scenario.

\subsection{Few-shot Semantic Segmentation}

Few-shot learning provides another way to reduce the reliance on human-labeled samples by recognizing novel classes with minimal support i.e., labeled samples.
The first work~\cite{shaban2017one} on few-shot segmentation proposed to adjust the final classifier with weights learned from support samples.

Recently, applying metric learning to classify images to the nearest class prototype extracted from support images has become popular in few-shot image classification~\cite{snell2017prototypical}. Dong et al.~\cite{dong2018few} were among the first to apply such a prototype-based framework to few-shot segmentation. Following their work, PANet~\cite{wang2019panet} introduced prototype alignment regularization by performing an additional query-to-support segmentation after the standard support-to-query segmentation. AMP~\cite{siam2019amp} learned the prototypes from multi-scale features and historical support samples, and Li et al.~\cite{li2021adaptive} proposed to consider multiple prototypes for each class.

Instead of using prototype as the anchor feature for classification, CRNet~\cite{liu2020crnet} and PFENet~\cite{tian2020prior} incorporated the prototypes into the query features and formulated the few-shot segmentation as a binary classification problem.

Our work is closely related to prototype-based few-shot segmentation methods such as PANet~\cite{wang2019panet}. However, our work differs in three aspects. First, we focus on applying few-shot learning to solve the challenging domain adaptation task. Second, most existing few-shot segmentation methods evaluated their performance on PASCAL VOC 2012 dataset~\cite{everingham2010pascal} and COCO dataset~\cite{lin2014microsoft} and mainly focused on recognizing ``things"~\cite{heitz2008learning} (objects such as car). In contrast, we consider a more general few-shot learning problem by further recognizing ``stuffs" (e.g., sky, road). Finally, existing few-shot segmentation methods consider a binary segmentation task for training, i.e., recognizing only one class for each step, which requires objects of that specific class to occupy a certain proportion of the training image. Such a requirement can be easily satisfied when images have low-resolution and contain only a few objects. However, compared with images of PASCAL and COCO, the urban images considered in this paper have significantly higher resolution and contain many more objects of different classes. For this, we developed a support image adaptive training strategy (details in Section \ref{sec_proto_semseg}) suited for few-shot segmentation on high-resolution images that contain arbitrary number of objects.

\subsection{Prototype Learning in Domain Adaptation}
Since the proposal of prototypical network~\cite{snell2017prototypical} for few-shot learning, incorporating prototype learning into domain adaptation has also received increasing attention in recent years. For example, Zhang et al~\cite{zhang2021prototypical} presented a prototype based denoise method to improve the quality of pseudo labels for the self-training framework. The motivation of this method can be summarized as: the pixel with noisy label is typically far from the class prototype corresponding to the same noisy label in the embedding space. Yue et al.~\cite{yue2021prototypical} developed a prototypical self-supervised learning framework by performing a cross-domain instance-prototype matching task. Different from these methods, in which the prototypes are usually used to facilitate the feature learning, our method focus on learn domain-invariant prototypes that will be directly used for classification. From this perspective, the work most related to ours is the transferable prototypical network (TPN) proposed by Pan et al.~\cite{pan2019transferrable}. However, TPN is an unsupervised domain adaptation framework and learns the prototypes through  minimizing domain discrepancy such as cross-domain prototype distances. In contrast, our method considers a few-shot supervised domain adaptation problem and the prototype learning are directly supervised by the classification loss. 

%is to multiply the soft pseudo label of a pixel with the similarity between the pixel and class prototypes since
%Thus, if a high probability is assigned to a class in a pseudo label, class prototype, such  are obtained by grouping the features of the same class indicated by current pseudo labels

\section{Preliminaries}
\label{sec_pre}
Since our work focuses on the domain adaptive semantic segmentation task and is developed from prototype-based few-shot semantic segmentation, we start by introducing commonly used formulations in these two tasks.

\subsection{Domain Adaptive Semantic Segmentation}
Domain adaptive semantic segmentation typically involves a source domain $S$ and a target domain $T$. The source domain contains a large number of images $I_s=\{I_j\}_{j=1}^{n_s}$ with dense annotations $Y_s=\{y_j\}_{j=1}^{n_s}$, where $n_s$ is the total number of images in $S$. The goal of domain adaptive semantic segmentation is to apply the knowledge learned from the source domain to the target dataset $I_t=\{I_j\}_{j=1}^{n_t}$ such that the requirement for annotations of target images is minimized.

\subsection{Few-shot Segmentation}
\label{sec_pre_fewshot}
Few-shot semantic segmentation is related to the few-shot learning which aims to recognize a set of novel test classes $C_{test}$ by learning models from a set of base training classes $C_{train}$ ($C_{train} \cap~C_{test} = \o$). Towards this goal, the samples of few-shot learning are typically divided into two sets, namely a support set $S_u$ (we added a subscript $u$ to differentiate it from above notation of source domain $S$)  and a query set $Q$. Given a query image $I_q$ from $Q$, the objective of few-shot semantic segmentation is to segment the objects of $C_{test}$ in $I_q$ with only a few labeled images from the support set.

Since a query image may contain more than one class of $C_{test}$, existing few-shot semantic segmentation tasks can be generalized to a $N$ way $K$ shot segmentation problem where $N$ represents the number of classes that are to be simultaneously recognized in a single segmentation process, and $K$ is the number of labeled images of each class of $N$.

Note that although the labeled samples for each test class should be limited to a small number $K$, for a $K$-shot segmentation, the number of labeled samples for the base classes are  usually available in large-scale to facilitate model training. Thus, the $K$-shot labeled samples are usually randomly sampled from the training set in each training episode. Suppose that a training batch contains $N$ classes that belong to $C_{train}$,  we need to provide $N\times K$ support images for training. In other words, we need to theoretically process $B+NK$ ($B$ is the batch size) images in total for each training step, which can lead to a huge GPU memory requirement if $NK$ is large (e.g., $10\times5$). Moreover, $N$ may also differ for each training step. Therefore, current few-shot learning methods usually consider a binary segmentation (N=1) problem during training.

\begin{figure}[t]
\centering
\includegraphics[width=3in]{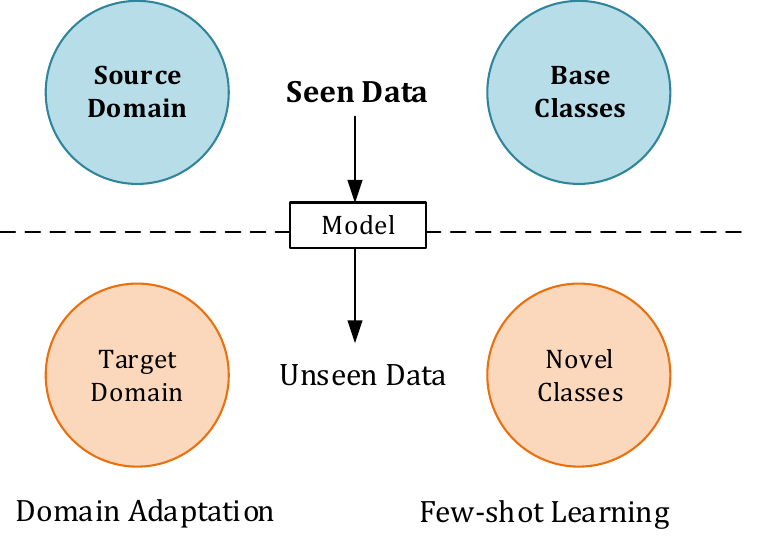}
\caption{Relationship between domain adaptation and few-shot learning. Both aim to recognize some unseen data with knowledge learned from a large amount of seen data}
\label{fig_motiv}
\end{figure}

\begin{figure*}[!ht]
\centering
\includegraphics[width=7.1in]{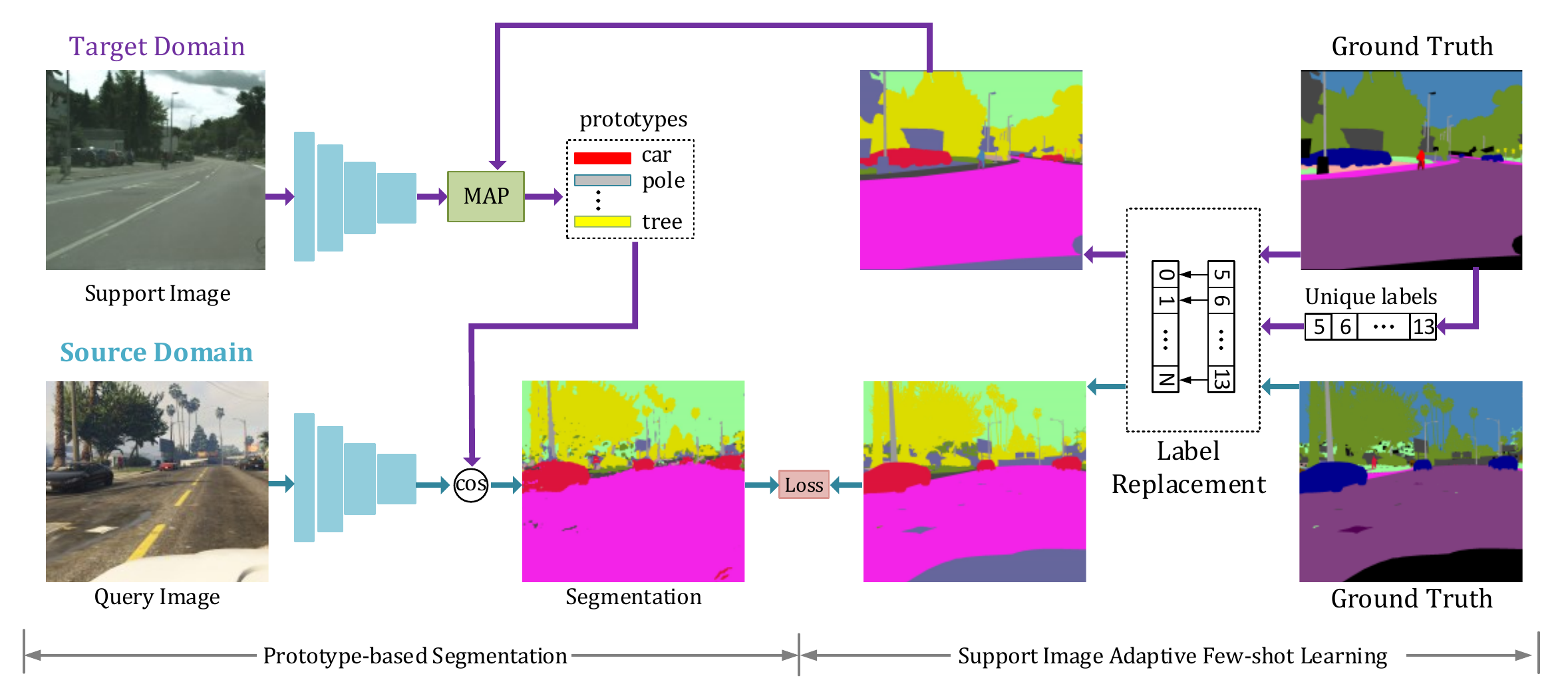}
\caption{Overview of our method. We propose to learn domain-invariant prototypes with a recent popular prototype-based few-shot learning framework. During training, we adopt few-shot annotated target images as the support set and treat all source images as the query set. In each training step, our model first embeds the support image  and query image  into semantic features using a Siamese Network. Then, a masked average pooling (MAP) operation is applied to the feature maps of support image to obtain class prototypes.
Finally, predictions over the pixels of query image are obtained by finding the nearest class prototype to each pixel.
%which are used to compute segmentation loss, are obtained by finding the class nearest to them reference to these prototypes.
To extend the prototype-based few-shot segmentation to high-resolution images containing arbitrary number of classes, we propose a support image adaptive training strategy in which the classes to be recognized as well as the number of support samples of them are fully determined by current support image.  For segmentation of a test image, we need only perform prototype-based segmentation with all the class prototypes that are pre-computed from these few-shot annotated target images.}
\label{fig_frame}
\end{figure*}

\subsection{Important Observation}
Figure \ref{fig_motiv} illustrates the relationship between domain adaptation and few-shot learning tasks. It can be observed that both the domain adaptive semantic segmentation and few-shot semantic segmentation aim to recognize some unseen data with knowledge learned from seen data. In particular, domain adaptive semantic segmentation aims to recognize unseen data distributions while few-shot semantic segmentation aims to recognize unseen classes. Thus, if we view all the classes presented in unseen distribution in the target domain as unseen classes to the source domain, the domain adaptive semantic segmentation can be naturally formulated into a few-shot semantic segmentation problem.

\section{Method}
\label{sec_method}
\subsection{Framework Overview}
\label{sec_method_frame}
Figure \ref{fig_frame} shows an overview of our method. Based on our hypothesis %assumption
that humans perform object recognition via
%a ``similarity comparing" operation,
``similarity comparisons'', we propose to predict the pixel labels by comparing its features to a number of domain-invariant class prototypes.

To learn domain-invariant prototypes, we collect a few annotated target images as the support set and treat all source images as the query set, following the common setting used for few-shot segmentation. For each training step, our model first embeds the support image (target domain) and query image (source domain) into semantic features using a Siamese Network. Then, a masked average pooling operation is applied to the feature maps of support image to obtain class prototypes. %Finally, prediction over the pixels of query image is performed by finding the class nearest to them reference to the prototypes.
Finally, predictions over the pixels of query image are obtained by finding the nearest class prototype to each pixel and assigning it the label of that prototype.

We use VGG-16~\cite{simonyan2014very} and adapt it into the DeepLab architecture for feature extraction. Specifically, we use the 5 convolutional blocks of VGG-16 as the backbone and remove the max pooling layer behind block 3 and block 4. To ensure the receptive field remains unchanged, the convolutions in block 4 and block 5 are replaced by dilated or atrous convolutions~\cite{chen2017deeplab} with dilation rates set to 2 and 4, respectively.

\subsection{Feature Refinement Module}
Intuitively, domain-invariant information is generally presented in high-level semantic concepts. For example, the semantic class label that we want the computer to recognize can be viewed as the highest-level domain-invariant information. Building on this insight, we further develop a feature refinement module (FRM) to encode as much semantic information as possible.

\begin{figure}[t]
\centering
\includegraphics[width=3.5in]{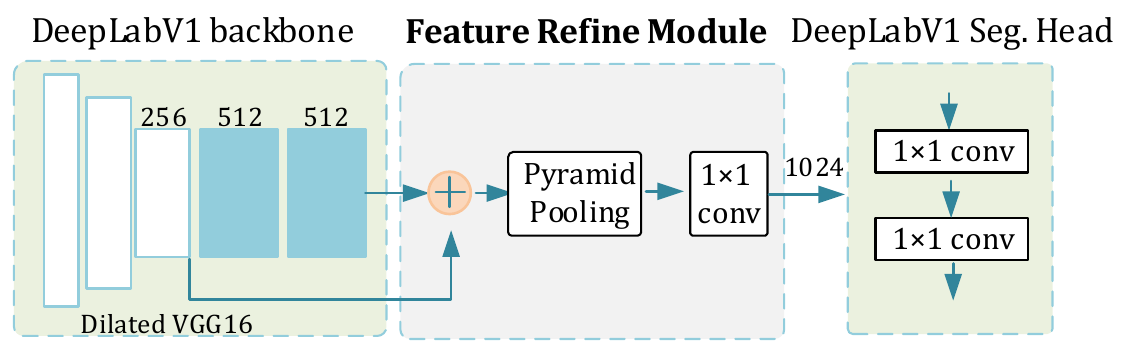}
\caption{Architecture of Feature Refinement Module (FRM).}
\label{fig_frm}
\end{figure}

As shown in the Fig.~\ref{fig_frm}, our FRM first concatenates the outputs of block 3 and block 5 of the backbone i.e., dilated VGG16. Next, we perform feature pyramid pooling, introduced in \cite{zhao2017pyramid}, to exploit context information. Finally, we adopt $1\times1$ convolution layers to transform the features to a lower dimensional (i.e. 1024) space.

\subsection{Construction of Support Set}
%As mentioned in Section \ref{sec_pre_fewshot}, %Ajmal: this just shows that you are repeating so no need to write this
For a $K$-shot semantic segmentation task, the $K$-shot labeled samples for training, i.e., the support set of each training episode, are usually randomly sampled from a large-scale dataset of base classes. However, since the support images for our training are sampled from target domain in which labels are difficult to obtain,  we must use a support set as small as possible. Below, we explain how we collect the support images from Cityscapes dataset~\cite{cordts2016cityscapes} that we used as target domain in this paper.

%For convenience, we call this array class occurrence array.
As shown in Algorithm \ref{alg_supportset_construction}, we first define an array of size $N_{class}$ (e.g., 19 in Cityscapes) with all elements initialized to zero. This array will be used to accumulate the number of occurrences of the 19 classes of Cityscapes during dataset collection. Next, we randomly sample an image from the training set ($N_{image}$) of Cityscapes and add the corresponding entries of the above array with $1$ if some classes are found in this image. Such accumulation of a specific class is ended if the occurrence of this class reaches to a predefined number $K_{shot}$ (similar to the $K$-shot). Only images that contribute novel accumulation to any class are then added to the support set. With this method, we finally collect 33 image-label pairs from the training set of Cityscapes for the support set by setting $K_{shot}=5$. For convenience, we will call this dataset Cityscapes-5Shot in the rest of the paper.

Note that the target domain is typically unlabeled in practice, which means we need to manually annotate a number of target images for support set construction in this case. To minimize the time costs on human-annotation, the number of annotation samples for each class can be provided at instance-level rather than image-level.

\begin{algorithm}
\caption{Construction of Support Set }
\begin{algorithmic}
\REQUIRE $ N_{image}, K_{Shot}, N_{class}$
\STATE $C_{occur}$ $\leftarrow$ create an array of zeros of
size $N_{class}$
\STATE $I_{list}$ $\leftarrow$ create an empty image list
\FOR{$i=0:1:N_{image}$}
\STATE{$l_{i}$ $\leftarrow$ get the unique labels contained in $i$ }
\STATE $n_i$ $\leftarrow$ get the image name of $i$
    \FOR{$c=0:1:N_{class}$}
        \IF{$C_{occur}[c] < K_{Shot}$ \& $c \in l_i$}
            \STATE $C_{occur}[c]+=1$
            \IF{$n_i \not\in I_{list}$}
                \STATE $I_{list}$ $\leftarrow$ {add $n_i$ to the $I_{list}$}
            \ENDIF

        \ENDIF
    \ENDFOR
\ENDFOR
\STATE Return ($I_{list}$)
\end{algorithmic}
\label{alg_supportset_construction}
\end{algorithm}

%\footnode{public available at}

%With the commonly used ``random resize and then random crop" sampling strategy, which first resize the input image to a random scale and then crop a patch from the resized image using random coordinates, the sampled training image can still reserve most parts of the input image.
\subsection{Prototype based Semantic Segmentation}
\label{sec_proto_semseg}

For a $N$-class semantic segmentation problem, the output at each pixel is a vector that consists of the probability of each class, which is usually computed by a softmax function as
\begin{equation}
b_c^{(h,w)} = \frac{\exp({\mathsf{w}_c^Tx_{h,w}})}
             {\sum_j^N \exp({\mathsf{w}_j^Tx_{h,w}})}
\label{eq_seg_prob},
\end{equation}
where $x_{h,w}$ is the feature vector of pixel at spatial location $(h,w)$ and $\mathsf{w}_j$ (we use a different font here to differentiate it from the $w$ used for location denotation) contains the classification parameters of class $j$. Then, the training loss over semantic segmentation is usually computed by averaging the cross entropy at each pixel:
\begin{equation}
\mathcal{L}_{seg}=\frac{1}{\mathcal{HW}} \sum_{h,w}-\log~ b_{\hat{c}}^{(h,w)}
\label{eq_loss_classicialseg},
\end{equation}
where $\mathcal{HW}$ denotes the size of input image,  $\hat{c}$ denotes the ground truth label.

Note that the value of $\mathsf{w}_c^Tx_{h,w}$ in Eq.~(\ref{eq_seg_prob}) is also called the score of class $c$. It essentially computes the distance between $x_{h,w}$ and $\mathsf{w}_c$ as an inner product. Since the $\mathsf{w}_c$ is fixed after model training, it can be viewed as a special prototype to some extent. Inspired by this observation, we propose to transform Eq.~(\ref{eq_seg_prob}) to a general prototype based version as
\begin{equation}
b_c^{(h,w)} = \frac
%{\exp \big( {\xi(dip_c, x_{h,w})}\big)} %Ajmal: need to change dip to a single symbol
%{\sum_j^N \exp\big({\xi(dip_j, x_{h,w})}\big)}
{\exp \big( {\xi(p_c, x_{h,w})}\big)} %Ajmal: need to change dip to a single symbol
{\sum_j^N \exp\big({\xi(p_j, x_{h,w})}\big)},
\label{eq_seg_ourprob}
\end{equation}
where $p_c$ is the domain-invariant prototype of class $c$, and $\xi$ is a similarity function. In this paper, we adopt the commonly used cosine similarity function
\begin{equation}
\xi(x_1,x_{2}) = \frac{x_1 \cdot x_{2}}{\max(\left\|x_1 \right\|_2 \cdot \left\|x_{2} \right\|_2, \epsilon) }.
\label{eq_cossimilar}
\end{equation}

%Ajmal: also in equations, use \max instead of max, \exp instead of exp

Note that Eq.~(\ref{eq_seg_ourprob}) requires the computation of the feature similarities with respect to $N$ class prototypes. We propose to compute these $N$ class prototypes by first computing prototypes separately from each image in Cityscapes-5shot dataset and then performing an averaging operation. Let $(I^{(j)}, Y^{(j)})$ be an image-label pair from the Cityscapes-5shot, our prototype extraction can be formally written as
\begin{equation}
p_c = \frac{\sum_{j=1}^{K} p_{c,j} \mathds{1}[c \in {\rm set}(Y^{(j)})]}
           {\sum_{j=1}^{K} \mathds{1}[c \in {\rm set}(Y^{(j)})]},
\label{eq_test_proto_compute}
\end{equation}
where $K$ is the total number of support images, $\mathds{1}[]$ is an indicator function that outputs 1 if the argument is true and 0 otherwise, set$()$ is a function that outputs the unique labels contained in a labeled image, and $p_{c,j}$ is the prototype of class $c$ extracted from $I^{(j)}$ with the masked average pooling function
\begin{equation}
p_{c,j}=\frac{\sum_{h,w}x_{h,w}^{(j)} \mathds{1}[Y_{h,w}^{(j)}=c]}
             {\sum_{h,w} \mathds{1}[Y_{h,w}^{(j)}=c]}.
\label{eq_mask_ave_pool}
\end{equation}

For the test stage, given the trained model, we can pre-compute the $p_{c}$  for all classes with Eq. (\ref{eq_test_proto_compute}) from the few-shot target images and save them to a local file. Then, we need only read these prototypes and applied them to Eq. (\ref{eq_seg_ourprob}) to perform semantic segmentation over test images.

For the training stage, however, $p_{c}$ need to be updated in each step, which could lead to a huge requirement on GPU memory if we consider the same number of classes as in testing/inference. In particular, we need to access all support images of the considered classes to extract the prototypes for each. The commonly used binary segmentation strategy (Section \ref{sec_pre_fewshot}) provides a solution to this problem, however, as detailed in Section \ref{sec_intro}, that training strategy is not suitable for high-resolution images containing arbitrary number of classes. Next, we will detail our support image adaptive training strategy for this problem.

\subsection{Support Image Adaptive Training}
\label{sec_method_trainingwith_prototypes}

%obtained by computing the intersection of these two sets as $C_{com}= C_{st} \cap ~C_{q}$
Given a support image from target domain and a query image from source domain, we propose to use the classes contained in the support image for model training. More formally, denoting the unique labels contained in the support image $I_{st}$ and the query image $I_{q}$ by a label set $C_{st}$ and $C_{q}$, respectively, we recognize only classes that belong to the $C_{st}$  in each training episode. Thus, supposing the number of elements of $C_{st}$ is $N_{class}$, we essentially consider a $N_{class}-$way few-shot segmentation problem. We note that $N_{class}$ could be different in different training steps. Moreover, the label sequence sorted from $C_{st}$ may be nonconsecutive (e.g, $C_{st}=\{1,7\}$), making it difficult to apply the commonly used cross entropy loss. To this end, we propose to replace the labels listed in $C_{st}$ with their entry indices, and all other labels with an invalid label 255. Given the above example $C_{st}=\{1,7\}$,  all pixels that belong to class 1 and 7 will be labeled 0 and 1, respectively. For the sake of clarity, we summarize our training sample generation in Algorithm \ref{alg_training_sample_ge}, where the $G_{st}$ and $G_{q}$ represent the ground truth label image of support image $I_{st}$ and query image $I_{q}$, respectively.

Notice that our few-shot learning framework does not require that each prototype class must be provided with $K-$shot samples during training.  Instead, the number of classes and their corresponding support samples are determined only by the current support image that is randomly sampled from the support set. This is reasonable since the number of annotated samples $K$ is not a constraint on the training stage but rather the test stage. Compared to the common binary segmentation based training, such a strategy has two important advantages. Firstly, it significantly reduces the number of support images required during training, resulting in a significant reduction in GPU memory consumption.
%Secondly, the binary segmentation usually limit to low-resolution images and adopt center cropping for training batch sampling,  which is necessary to ensure some parts of objects are sampled for loss computation.
%In contrast, since we consider only the classes after sampling,  our method can be applied to images without any constraints on both image resolution and number of objects.
Secondly, since we consider only the classes after sampling,  our method can be applied to images without any constraints on their resolution or the number of objects contained.

%Note that the case $C_{st} \cap ~C_{q}=\O$ is also likely to happen during sampling, which means all the pixel of $I_q$ will be labeled to the invalid label. Experiments show that our

%Thus, we first randomly sample a support image from the Cityscapes-5Shot, and then randomly sample a query image from GTA5 in a ``while'' circle until $|C_{com}| \ge 2$. In other words, we consider at least 2 classes in each training episode.

%In this case, the training batch consists of $B+K$ samples of this class, where $B$ and $K$ is the number of query images and support images, respectively.
%are typically smaller than 19, i.e., the total number of classes.

\begin{algorithm}
\caption{Support Image Adaptive Training}
\begin{algorithmic}
\REQUIRE $ I_{st}, G_{st}, I_{q}, G_{q}$
\STATE {$C_{st}$ $\leftarrow$  get the unique labels contained in $G_{st}$}
\STATE $C_{st}$ $\leftarrow$ delete the invalid label in $C_{st}$
\STATE {$C_{q}$ $\leftarrow$ get the unique labels contained in $G_{q}$}

\FOR{$i \in C_q$}
    \IF{$ i \not \in C_{st}$}
        \STATE $G_{q}$ $\leftarrow$ replace label $i$ in $G_{q}$ with the invalid label
    \ENDIF
\ENDFOR

\FOR{$i \in C_{st}$}
    \STATE $e_i$ $\leftarrow$ get the entry index of $i$ in $C_{st}$
    \STATE $G_{st}, G_{q}$ $\leftarrow$  replace label $i$ in $G_{st}$ and $G_{q}$ with $e_i$
\ENDFOR

\STATE Return $I_{st}, G_{st}, I_{q}, G_{q}$
\end{algorithmic}
\label{alg_training_sample_ge}
\end{algorithm}

\subsection{Loss Function}
Similar to most prototype-based few-shot segmentation methods, our basic few-shot domain adaptation framework requires only the annotations of support images for prototype extraction and the prototype-based segmentation is supervised by the dense annotations of query images. While the support image and query image in our method come from different domains and our goal is to learn domain-invariant prototypes, we further apply the  annotations of support images to supervise the prototype-based segmentation on themselves. Thus, the final loss for our model training is composed of two parts: the segmentation loss on query images $\mathcal{L}_{seg}^{Q}$ and the segmentation loss on support images $\mathcal{L}_{seg}^{ST}$. Both are obtained by computing the cross entropy loss via Eq.~(\ref{eq_loss_classicialseg}). Given that the support images are few, we treat the segmentation loss on support images as an auxiliary loss and multiply it with a balancing factor $\alpha$ to avoid overfitting to the few support images. Our final loss can be formally written as
\begin{equation}
\mathcal{L} = \mathcal{L}_{seg}^{Q} + \alpha \mathcal{L}_{seg}^{ST}.
\label{eq_final_loss}
\end{equation}

\begin{table*}[ht]
\renewcommand\arraystretch{1.5}
\caption{Performance of our Baseline on GTA5-to-Cityscapes adaptation. Note that we use only  $1/10$ source images in this experiment. The ``stloss'' represents the auxiliary segmentation loss on support image, and the followed number is the balancing factor .}

\centering

\begin{tabular}{l  p{0.25cm} p{0.25cm} p{0.25cm} p{0.25cm} p{0.25cm} p{0.25cm} p{0.25cm} p{0.25cm} p{0.25cm} p{0.25cm} p{0.25cm} p{0.25cm} p{0.25cm} p{0.25cm} p{0.25cm} p{0.25cm} p{0.25cm} p{0.25cm} p{0.25cm} p{0.25cm} }
\hline

Model              &\rotatebox{90}{road}      &\rotatebox{90} {side.}   &\rotatebox{90} {build.}     &\rotatebox{90}{wall}     & \rotatebox{90}{fence}
&\rotatebox{90}{pole}     & \rotatebox{90}{light}  &\rotatebox{90}{sign}     &\rotatebox{90}{vege.} &\rotatebox{90}{terr.}   & \rotatebox{90}{sky }      &\rotatebox{90}{person}    &\rotatebox{90}{rider}  &\rotatebox{90}{car}    & \rotatebox{90}{truck}    &\rotatebox{90}{bus }       &\rotatebox{90}{train  }   &\rotatebox{90}{motorcy.}   &\rotatebox{90}{bike} &\rotatebox{90}{\textbf{mIoU}} \\
\hline
Source Only      & 60.3   & 21.2  &61.9    & 3.7 & 13.1   &21.6&20.7 &5.7 &77.3  &25.6
                    & 22.7 & 32.2  &0.0     &40.3 & 5.2 &1.7 & 0.0 &0.0  &0.0   &\textbf{21.8} \\
Few-shot Only (5 shot)      & 88.4   & 42.5  &76.8    & 9.9 & 10.8   &12.7&0.3 &27.7 &83.5  &26.6
                    & 68.2 & 41.6  &0.0     &71.6 & 3.8 &2.1 & 0.0 &0.0  &26.2   &\textbf{31.2} \\
Few-shot Only (1 shot)     & 85.1  & 24.7 &68.8   & 4.2 & 0.0   &0.0&0.0 &0.0 &82.7  &16.6
                    & 41.8 & 16.2 &0.0     &55.9 & 0.0 &0.0 & 0.0 &0.0  &0.8   &\textbf{20.9} \\
\hline
FSDA    & 82.8  & 37.6  &69.3   &13.6 & 15.2   &23.3&12.3 &31.6 &78.3  &22.9
                    & 70.5 & 38.6  &5.7    &63.6 & 4.4 &6.7 & 1.2 &2.6  &15.6   &\textbf{31.4} \\
FSDA+stloss~(0.1)      & 91.7  & 51.0 &78.8   &17.5 & 12.5   &29.0&22.2 &40.0 &83.3  &26.2
                    & 76.7& 39.1  &6.5    &73.9 & 5.2 &12.6 & 1.9 &4.1  &14.6   &\textbf{36.2} \\
FSDA+stloss~(0.2)      & 91.4  & 52.5  &79.8   &18.0 & 14.9   &30.8&22.1 &38.4 &84.0  &28.9
                    & 80.0 & 35.6  &7.1    &73.4 & 4.8 &13.2 & 1.8 &4.6  &15.8   &\textbf{36.7} \\
FSDA+stloss~(0.4)     & 91.6  & 49.2  &79.3   &15.3 & 13.2  &28.5&23.7 &40.1 &84.4  &32.2
                    & 74.7 & 37.4  &7.1    &75.3 & 5.7 &11.3 & 2.2 &4.7  &16.6   &\textbf{36.5} \\
FSDA+stloss~(0.6)      & 91.9  & 53.9  &79.5   &14.3 & 14.7  &29.2&24.0 &39.2 &84.2  &29.0
                    & 72.4 & 39.2  &6.8   &75.2 & 5.3 &10.6 & 2.1 &5.4  &16.2   &\textbf{36.5} \\

FSDA (unseen)+stloss~(0.2)      & 92.6  & 54.0  &79.0  &14.5& 16.3   &29.2&26.1 &40.9 &84.3  &29.3
                    & 80.4 & 34.2  &8.9    &72.4 & 7.2 &7.1 & 2.6&4.2  &17.0   &\textbf{36.9} \\
\hline
FSDA (1 shot)     & 85.8  & 37.4  &70.8   &13.7& 12.9   &23.9&18.0 &28.9 &78.8  &24.6
                    & 51.9 & 39.5  &7.5    &71.2 & 5.1 &7.8 & 1.8&4.7  &9.9   &\textbf{31.3} \\
FSDA (1 shot)+stloss (0.2)      & 88.7  & 35.7  &76.9   &16.6& 10.5   &26.4&25.6 &33.2 &83.4  &28.0
                    & 53.5 & 49.5  &6.1    &74.0 & 5.2 &6.7 & 2.1&4.8  &15.5   &\textbf{33.8} \\

\hline

\end{tabular}
\label{table_results_baseline}
\end{table*}

\section{Experiments}
\label{sec_exp}
%Similar to many existing domain adaptive semantic segmentation methods,
Following common practice in domain adaptive semantic segmentation research, we first validate the effectiveness of our method by conducting experiments on  GTA5-to-Cityscapes adaptation task.  The Cityscapes~\cite{cordts2016cityscapes}  dataset contains 2,975, 500, and 1,525 densely annotated urban images for training, validation, and testing, respectively. The images are collected from 50 European cities. Note that the annotations of 1,525 test images are not publicly available.  The GTA5 dataset~\cite{richter2016playing} contains 24,966 synthetic images from a well-known computer game named Grand Theft Auto. It shares 19 common labels with the Cityscapes dataset.

\subsection{Implementation Details}
Since the GTA5 consists of  24,966 images with resolution up to $1914\times1052$, training over the GTA5 is usually time-consuming. Hence, the GTA5 contributors divided the data into 10 parts. For our ablation studies, we use only the first part which contains 2,500 images. After that, we train the final model over the entire GTA5 dataset for comparison to state-of-the-art methods.

%Similar to most semantic segmentation methods,
We initialized the VGG16 backbone with ImageNet pre-trained weights. Next, we divided the VGG16 into five blocks according to the feature resolution and fixed
the weights of  %check
the first three blocks when adapting it to the segmentation task. For convenience, we will refer to our few-shot domain adaptation method as FSDA  in the rest of the paper.

\subsection{Experimental Settings}
\label{sec:experimentSettings}
%\textbf{Training policy:}
%For the sake of clarity, we detailed the main setting of our training policy in the Table \ref{table_expsetting}.
We implemented and trained our deep model in PyTorch~\cite{paszke2019pytorch}.
We trained our model for 50 epochs on $1024\times 1024$ images (batch size 1) using the SGD optimizer with a base learning rate of 0.001 and momentum 0.9. The learning rate was updated with the commonly used ``poly" policy~\cite{chen2017deeplab}. For data augmentation, we used random scaling (0.9 to 1.1), random horizontal flip and random rotation ($-10^{\circ}$ to $10^{\circ}$). Our  training batch sampling follows the commonly used ``scaling-and-cropping" strategy which first resizes a training image to a random scale and then crops an image patch equal to the training size from the scaled image.  All experiments were conducted on a computer equipped with a Titan X (12GB) GPU and 32G RAM.
%\textbf{Evaluation Metrics:}
We adopt the commonly used mean intersection over union (mIoU) as the major metric for model comparison.
%\textbf{Experiment environments:}

\subsection{Comparison to Baselines}

We first implemented the most straightforward adaptation strategy, i.e., directly applying model trained on GTA5 to Cityscapes, as baseline and achieved 21.8\% mIoU (see Table \ref{table_results_baseline}).
Since our method utilized few-shot annotated target images for domain adaptation, we further evaluated another baseline named few-shot only in which the model was trained on the few-shot (33) support images of Cityscapes.
%According to Table \ref{table_expsetting},
Our few-shot domain adaptation is trained for 50 epochs in the following experiments. Each training image is associated with a support image from the 33 few-shot images in this case. In other words, the few-shot training set would be traversed for $(50\times 2500) /33 \approx 3788$ epochs in our FSDA framework.  Thus, for a fair comparison, we should train the model for 3788 epochs for the few-shot baseline. However, the trained model may easily overfit to the few-shot training images given such a large number of training epochs. On the other hand, we found that the few-shot baseline starts to converge at about 180 epochs. Thus, we set the training epochs to 200 for the few-shot baseline, which yielded 31.2\% mIoU on the validation set of Cityscapes.

As shown in Table \ref{table_results_baseline}, compared to using the source only, our FSDA framework obtained a performance gain of 9.6 percentage points (31.4\% mIoU), showing the effectiveness of our few-shot adaptation framework. Compared to the few-shot only baseline, our FSDA obtained only 0.2 percentage point performance gain. Nevertheless,  our method can  segment  classes (e.g., rider, train and motorcycle) that are generally missed by the few-shot only baseline. Later in Section \ref{sec_exp_oneshot}, we  show that our method can boost this gain significantly when fewer support images are used. Moreover, we used these few-shot annotations only for prototype extraction in our basic FSDA framework. Our following experiments show that we can obtain more gains when these annotations are further used to directly supervise the segmentation model.

\begin{figure*}[!ht]
\centering
\includegraphics[width=7.1in]{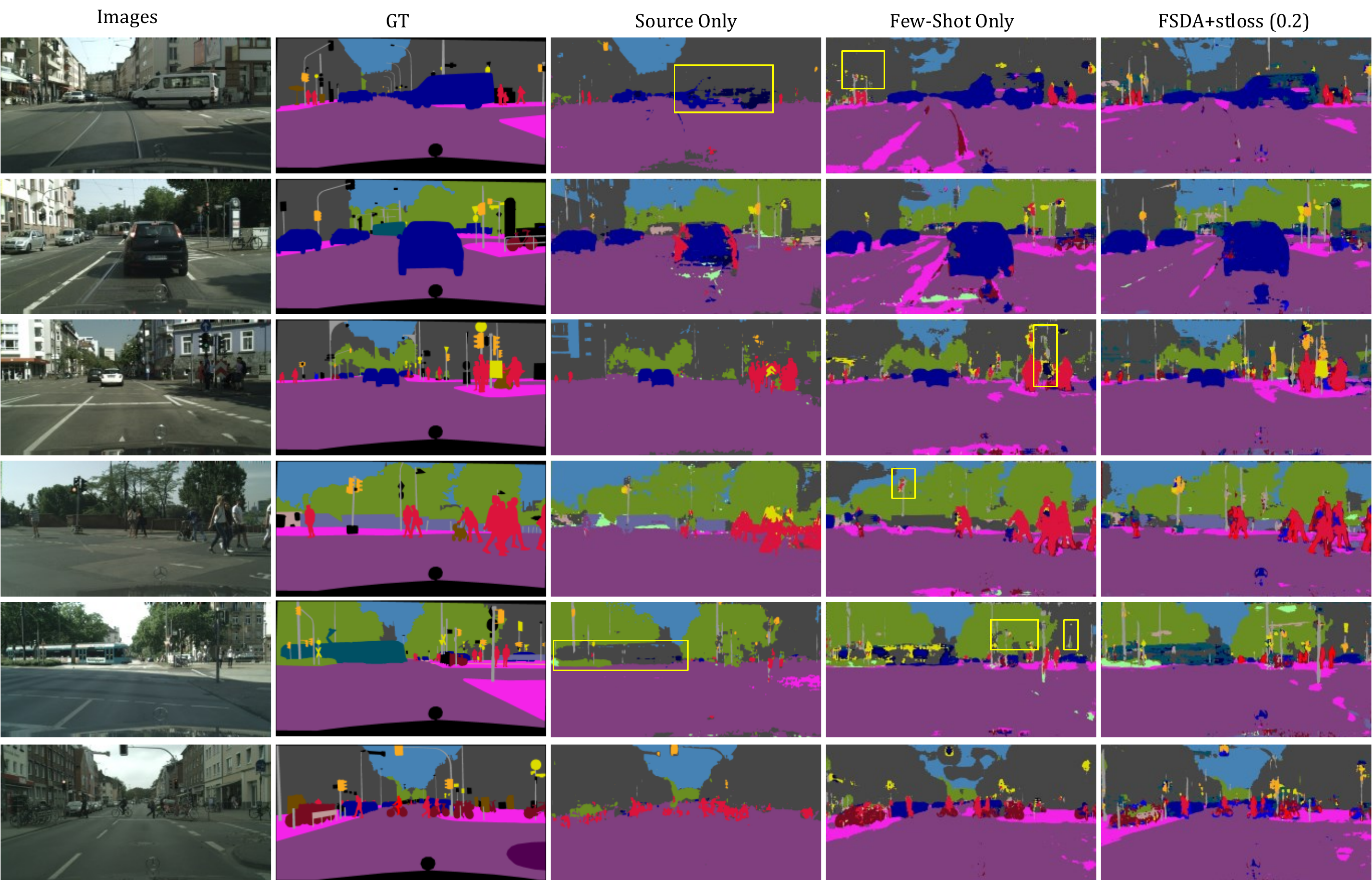}
\vspace{-3mm}
\caption{Qualitative results on GTA5-to-Cityscapes adaptation. The source only baseline has a bias towards stuff classes that occupy a large continuous area, while the few-shot only baseline cannot recognize small objects well due to the limitation on the total number of training pixels.}
\label{fig_visualization}
\end{figure*}

\subsection{Ablation Study on Loss}
Our basic FSDA method employed only $\mathcal{L}_{seg}^{Q}$  for model training. To validate how the auxiliary segmentation loss $\mathcal{L}_{seg}^{ST}$ in Eq.~(\ref{eq_final_loss}) influences the training, we tested different values of the balancing factor.

We first tested a small balancing factor of 0.1 and achieved 36.2\% mIoU (see Table \ref{table_results_baseline}), obtaining a performance gain of 4.8\% mIoU over our basic FSDA framework. The performance further improved to 36.7\% mIoU when the balancing factor was increased to 0.2. Since these performances are on the target domain, this can be well explained because $\mathcal{L}_{seg}^{ST}$ directly utilized the supervision signal from the target domain.  However, we did not obtain further gain but rather a reduction in mIoU when we further increased the balancing factor, i.e., letting the supervision from target domain play a much greater role in model training. In particular, the performance in terms of mIoU decreased from 36.7\% to 36.5\% when the balancing factor was increased from 0.2 to 0.4. This is mainly because the annotated target images we used were very few. Thus, increasing the weight of $\mathcal{L}_{seg}^{ST}$ would also promote the prototype learning to overfit to these few-shot annotations.

With our best performing FSDA framework, we show qualitative comparisons between our method and baselines in Fig. \ref{fig_visualization}. It can be observed that the source only baseline has a bias towards stuff classes that usually occupy a large continuous area such as wall and road classes. For example, almost the whole car (center-right) in the first image and the whole bus in the fifth image are classified into wall by the source only baseline. This bias was greatly reduced by our framework. Compared with the few-shot baseline, our method performed much better on small objects such as traffic light, traffic sign, and pole. This is mainly because pixel-wise loss based training can be easily dominated by large objects especially when the total training samples are few. %presented in a small number.

 \subsection{Generalization to Unseen Support Image}
We used the same support images for training and testing in the above experiments. In this experiment, we generate three groups of support images that are different from those 33 images used for training, which can be easily implemented by applying a random shuffle operation with different seeds to the training set of Cityscapes before employing Algorithm \ref{alg_supportset_construction}.  However, this is not recommended as this would increase the requirement for annotations on the target domain. The main purpose of this experiment is to simply verify whether our FSDA can generalize to unseen support images. Thus, these three groups support images were only used for testing. As shown in the third last row of Table \ref{table_results_baseline}, our FSDA achieved an average 36.9\% mIoU with these three group unseen support images, which is similar to the performance achieved with the seen support images.

\subsection{One Shot Domain adaptation}
\label{sec_exp_oneshot}
We used the Cityscapes-5Shot dataset as the support set in the above experiments. To further validate whether our method can work with fewer labeled target images, we considered a one-shot domain adaptation problem. To this end, we generated a Cityscapes-1Shot dataset that contains only 7 annotated images for training using  Algorithm \ref{alg_supportset_construction}.

Similarly, we first implemented the few-shot only baseline with Cityscapes-1Shot and achieved 20.9\% mIoU i.e. a 10.3 percentage point drop compared with using Cityscapes-5Shot. This is mainly due to the difficultly of learning effective features for ``things" objects (e.g. car, pole) from only one-shot samples, especially when ``stuff" objects (e.g. road) occupy larger proportions of the images. As shown in Table \ref{table_results_baseline},  almost all ``things" objects except the car could not be correctly segmented by the few-shot only baseline when only one-shot support images were used. In contrast, our method still achieved  a relatively steady performance (33.8\% mIoU) compared with the 36.7\% mIoU achieved in 5-shot experiments. Moreover, our method outperformed the source only baseline and few-shot only baseline by 12 and 12.9 percentage points, respectively.

%Ajmal: your performance is not the best so you need to inculde something other than year to make your point e.g. target domain samples
%
\begin{table*}[ht]
\renewcommand\arraystretch{1.4}
\caption{Performance comparison with state-of-the-arts on GTA5-to-Cityscapes adaptation.}
\vspace{-3mm}
\centering

\begin{tabular}{l  p{0.25cm} p{0.25cm} p{0.25cm} p{0.25cm} p{0.25cm} p{0.25cm} p{0.25cm} p{0.25cm} p{0.25cm} p{0.25cm} p{0.25cm} p{0.25cm} p{0.25cm} p{0.25cm} p{0.25cm} p{0.25cm} p{0.25cm} p{0.25cm} p{0.25cm} p{0.25cm} }
\hline

Model             &\rotatebox{90}{road}      &\rotatebox{90} {side.}   &\rotatebox{90} {build.}     &\rotatebox{90}{wall}     & \rotatebox{90}{fence}
&\rotatebox{90}{pole}     & \rotatebox{90}{light}  &\rotatebox{90}{sign}     &\rotatebox{90}{vege.} &\rotatebox{90}{terr.}   & \rotatebox{90}{sky }      &\rotatebox{90}{person}    &\rotatebox{90}{rider}  &\rotatebox{90}{car}    & \rotatebox{90}{truck}    &\rotatebox{90}{bus }       &\rotatebox{90}{train  }   &\rotatebox{90}{motorcy.}   &\rotatebox{90}{bike} &\rotatebox{90}{\textbf{mIoU}} \\
\hline

% &\multicolumn{21}{c}{Self-training based methods}\\
%\hline
FCN wild~\cite{hoffman2016fcns}         & 70.4    & 32.4   &62.1       & 14.9  & 5.4     &10.9 &  14.2  & 2.7  &79.2     &21.3 & 64.6    & 44.1   &4.2       &70.4  &  8.0    & 7.3 & 0.0   &3.5  & 0.0     &27.1  \\

CDA~\cite{zhang2017curriculum}      & 74.9   & 22.0  &71.7     & 6.0 & 11.9   &8.4 &  16.3  &11.1 &75.7    &13.3                & 66.5 & 38.0   &9.3       &55.2 &  18.8    &18.9 & 0.0  &16.8  &14.6    &28.9  \\

MCD~\cite{saito2018maximum}     & 86.4   & 8.5  &76.1      & 18.6 & 9.7   &14.9 &  7.8  &0.6 &82.8    &32.7   & 71.4  & 25.2   &1.1       &76.3 &  16.1    & 17.1 & 1.4  &0.2  & 0.0     &28.8  \\

I2I~\cite{murez2018image}      & 85.3  & 38.0  &71.3      & 18.6 & 16.0   &18.7 &  12.0  &4.5 &72.0   &\textbf{43.4}   & 63.7  & 43.1   &3.3       &76.7 &  14.4   & 12.8 & 0.3 &9.8 & 0.6    &31.8  \\

CyCADA ~\cite{hoffman2018cycada}       & 85.2    & 37.2 &76.5       & 21.8 & 15.0   &23.8 & 22.9  & 21.5 &80.5    &31.3    & 60.7  & 50.5   &9.0       &76.9 &  17.1    & 28.2 & 4.5 &9.8  & 0.0     &35.4  \\
CBST~\cite{zou2018unsupervised}      & 90.4   & 50.8  &72.0     & 18.3 & 9.5   &27.2&  28.6  &14.1 &82.4    &25.1                    & 70.8 & 42.6   &14.5      &76.9 &  5.9   &12.5 & 1.2 &14.0  &28.6   &36.1\\

Advent~\cite{vu2019advent}       & 86.9   & 28.7 &78.7      & 28.5 & 25.2  &17.1 &  20.3  & 10.9  &80.0     &26.4 & 70.2    & 47.1   &8.4     &81.5  &  26.0   & 17.2 & 18.9 &11.7  & 1.6     &36.1  \\

PyCDA~\cite{lian2019constructing}       & 86.7   & 24.8  &80.9       & 21.4  & 27.3   &30.2 &  26.6  & 21.1  &\textbf{86.6}     &28.9 & 58.8    & 53.2   &17.9      &80.4  &  18.8   & 22.4 & 4.1  &9.7  & 6.2     &37.2  \\

DPR~\cite{tsai2019domain}      & 87.3   & 35.7  &79.5       & 32.0 & 14.5   &21.5 &  24.8  & 13.7  &80.4     &32.0 & 70.5    & 50.5  &16.9      &81.0  & 20.8   & 28.1 & 4.1  &15.5  & 4.1     &37.5  \\

CLAN~\cite{luo2021category}       & 90.4  & 40.2 &80.6   & 25.1 & 21.8   &27.6 &  24.2  & 19.6 &83.1    &33.9 & 74.1  & 47.7   &9.5    &83.9 &  27.0  & 27.1 & 3.4 &17.5 & 0.9  &38.8  \\
\hline
PIT~\cite{lv2020cross}       & 86.2  & 35.0  &82.1     & 31.1 & 22.1   &23.2 &  29.4  & 28.5 &79.3    &31.8 & 81.9    & 52.1   &23.2     &80.4  &  29.5   & 26.9 & 30.7  &20.5  & 1.2     &41.8  \\

FDA~\cite{yang2020fda}      & 86.1  & 35.1  &80.6     & 30.8 & 20.4   &27.5 &  30.0  & 26.0 &82.1    &30.3 & 73.6    & 52.5   &21.7    &81.7 &  24.0   & 30.5 & 29.9  &14.6  & 24.0     &42.2  \\
FADA~\cite{wang2020classes}       & \textbf{92.3}  & 51.1 &83.7   & 33.1 & 29.1   &28.5 &  28.0  & 21.0 &82.6    &32.6 & 85.3   & 55.2   &\textbf{28.8}    &83.5 &  24.4   & 37.4 & 0.0  &21.1 & 15.2    &43.8  \\

LDR~\cite{yang2020label}     & 90.1  & 41.2 &82.2    & 30.3 & 21.3   &18.3 &  33.5  & 23.0 &84.1    &37.5 & 81.4   & 54.2   &24.3    &83.0 &  27.6   & 32.0 & 8.1  &\textbf{29.7}  & 26.9    &43.6  \\
CD-AM~\cite{yang2021context}       & 90.1  & 46.7 &82.7   & \textbf{34.2} & 25.3   &21.3 &  33.0  & 22.0 &84.4    &41.4 & 78.9   & 55.5   &25.8    &83.1 &  24.9   & 31.4 & 20.6  &25.2 & 27.8    &44.9  \\
SA-I2I~\cite{musto2020semantically}       & 91.1  & 46.4 &82.9   & 33.2 & 27.9   &20.6 &  29.0  & 28.2 &84.5    &40.9 & 82.3  & 52.4   &24.4    &81.2 &  21.8  & 44.8 & \textbf{31.5} &26.5 & 33.7   &46.5  \\
SAC~\cite{araslanov2021self}      & 90.0  & \textbf{53.1} &\textbf{86.2}   & 33.8 & \textbf{32.7}   &\textbf{38.2} &  \textbf{46.0} & 40.3 &84.2    &26.4& \textbf{88.4} & \textbf{65.8}   &28.0   &\textbf{85.6} &  \textbf{40.6}  & \textbf{52.9} & 17.3 &13.7 & 23.8  &\textbf{49.9}  \\

\hline

Ours (1shot)   & 88.9  & 39.0  &78.9   & 19.2 & 14.9  &27.1& 24.7 &35.4 &84.6   &34.5
                    & 66.4 & 51.5  &4.2    &77.6 & 5.3  &3.5 & 1.7 &5.0 &8.0 &\textbf{35.3} \\
Ours (5shot)    & 90.6  & 47.0  &80.1   & 15.4 & 15.6   &29.3&  30.7 &\textbf{46.1} &82.4   &26.4
                    & 76.4 & 51.9  &12.3    &76.3 & 7.1  &9.7 & 1.2 &7.6  &\textbf{42.0} &39.4 \\
\hline

\end{tabular}
\label{table_sota}
\end{table*}

\subsection{Comparison With State-of-The-Art}
\label{sec_exp_comp_sota}
For comparison to representative methods published in recent years, we trained our FSDA on the whole GTA5 dataset for 10 epochs using the same training policies
% in Table \ref{table_expsetting},
listed in Section \ref{sec:experimentSettings}. 
Table \ref{table_sota} shows detailed comparative results. The proposed FSDA achieved
39.4\% mIoU on Cityscapes validation set outperforming many methods (e.g., FCN wild, CDA, MCD) that were published before 2020. When compared with state-of-the-art results obtained in past two years, our method lost its competitiveness. For example, the SAC method achieved close to 50\% mIoU with VGG16. Nevertheless, it is noteworthy that the main purpose of this paper is not to improve the state-of-the-art performance achieved by current self-training based methods and adversarial training based methods, but rather to provide a simple generic alternative for the domain adaptation task. Our one-stage training method is much easier to implement compared to existing methods since it does not require either multiple rounds self-training on large-scale target images or optimizing adversarial losses.  Moreover, while these compared methods only focus on a standard domain adaptation task, our method can be generalized to many variants of both domain adaptation task and few-shot learning tasks as follows.

\noindent \textbf{Open Set Domain adaptation (OSDA)}:  Denoting the label set considered in source domain and target domain by $C_S$ and $C_T$, respectively, there are many possible variants for domain adaptation. Most existing methods consider a standard domain adaptation problem where $C_S = C_T$. However, less attention has been paid to the OSDA problem, where $C_S \neq  C_T ~\& ~C_S \cap C_T = C_{com} \neq \emptyset$, i.e. there are a few private classes in target domain or both. It can be observed that the OSDA needs to recognize some novel classes in target domain, which shares a common character with the few-shot learning task. In other words, the OSDA can be naturally extended to a few-shot learning task with domain shifts if some annotated samples are available for each private class of the target domain.  Thus, our FSDA framework, developed from prototype-based few-shot learning, can also be easily applied to solve the OSDA problem. In particular, we can use classes that are contained in both the target and source domain, i.e., $C_{com}$, as the base classes to train the network for domain-invariant prototype extraction. Then, for both base classes and novel classes in target domain, we need only perform prototype-based segmentation using Eq.~(\ref{eq_seg_ourprob}) with class prototypes extracted from their few-shot annotated samples. We perform OSDA experiments in Section \ref{sec_synthia2cityscapes}.

\noindent \textbf{Multi-Source Domain adaptation (MSDA)}:  The MSDA task considers a case that there are multiple source domains presented in different distributions for knowledge transfer. Similar to the standard domain adaptation, the basic MSDA assumes that all domains share the same label set, i.e., $\forall{i},  C_S^{(i)}=C_T$, where $i$ is the index of source domain.
The basic MSDA can be evolved into a multi-source open set domain adaptation (MS-OSDA) problem if there are a few classes contained only in the target domain.  Since the core idea of our FSDA is to learn domain-invariant prototypes, i.e., we do not care about the domain-wise labels of images and need only to merge all source domains to a single domain for the basic MSDA problem. With this setting,  the MS-OSDA problem can also be solved with the method for OSDA mentioned above.

\noindent \textbf{Generalized Few-Shot Learning:} Compared to state-of-the-art domain adaptation, another important advantage of our method is that it can also be used to solve a generalized few-shot learning problem without constraints on the image resolution and the number of classes (see Section \ref{sec_method_trainingwith_prototypes}). %Ajmal: check the Section reference %\ref{sec_method_trainingwith_prototyp}).

%According to the denotation of Section \ref{sec_pre_fewshot}, a few-shot learning problem aims to recognize a number of classes $C_{test}$ on condition that there are only $K-$shot annotated samples for each class. Towards this goal, current few-shot learning usually learn to perform this challenging task on a number of basic classes $C_{train}$ that associated with a large amount of annotated samples.

%REF TO cvpr2021 PIXELMATCH

\begin{table*}[ht]
\renewcommand\arraystretch{1.4}
\caption{Performance comparison with state-of-the-art on SYNTHIA-to-Cityscapes adaptation.}%, GO for general objects.}
\vspace{-3mm}
\centering

\begin{tabular}{l  p{0.25cm} p{0.25cm} p{0.25cm} p{0.25cm} p{0.25cm} p{0.25cm} p{0.25cm} p{0.25cm} p{0.25cm} p{0.25cm} p{0.25cm} p{0.25cm} p{0.25cm} p{0.25cm} p{0.25cm} p{0.25cm} p{0.25cm} p{0.25cm} p{0.25cm} p{0.25cm} p{0.25cm} p{0.25cm}}
\hline

Model               &\rotatebox{90}{road}      &\rotatebox{90} {side.}   &\rotatebox{90} {build.}     &\rotatebox{90}{wall}     & \rotatebox{90}{fence}
&\rotatebox{90}{pole}     & \rotatebox{90}{light}  &\rotatebox{90}{sign}     &\rotatebox{90}{vege.} &\rotatebox{90}{terr.}   & \rotatebox{90}{sky }      &\rotatebox{90}{person}    &\rotatebox{90}{rider}  &\rotatebox{90}{car}    & \rotatebox{90}{truck}    &\rotatebox{90}{bus }       &\rotatebox{90}{train  }   &\rotatebox{90}{motorcy.}   &\rotatebox{90}{bike} &\rotatebox{90}{\textbf{mIoU16}} &\rotatebox{90}{\textbf{mIoU13}} &\rotatebox{90}{\textbf{mIoU19}}\\
\hline

% &\multicolumn{21}{c}{Self-training based methods}\\
%\hline
%FCN wild~\cite{hoffman2016fcns}  &       & 70.4    & 32.4   &62.1       & 14.9  & 5.4     &10.9 &  14.2  & 2.7  &79.2     &21.3 & 64.6    & 44.1   &4.2       &70.4  &  8.0    & 7.3 & 0.0   &3.5  & 0.0     &\textbf{27.1}  \\

CDA~\cite{zhang2017curriculum}     & 65.2   & 26.1  &74.9     & 0.1 & 0.5   &10.7 &  3.7  &3.0 &76.1    &-                & 70.6 & 47.1  &8.2      &43.2 &  -    &20.7 & -  &0.7  &13.1   &28.9  &-&-\\

%MCD~\cite{saito2018maximum}   &CVPR2018    & 84.8   & 43.6  &79.0    & 3.9 & 11.9   &8.4 &  16.3  &11.1 &75.7    &-                & 66.5 & 38.0   &9.3       &55.2 &  -    &18.9 & -  &16.8  &14.6    &\textbf{28.9}\\

%CyCADA ~\cite{hoffman2018cycada} &ICML2018      & 85.2    & 37.2 &76.5       & 21.8 & 15.0   &23.8 & 22.9  & 21.5 &80.5    &31.3    & 60.7  & 50.5   &9.0       &76.9 &  17.1    & 28.2 & 4.5 &9.8  & 0.0     &\textbf{35.4}  \\

Advent~\cite{vu2019advent}      & 67.9   & 29.4  &71.9     & 6.3 & 0.3   &19.9 &  0.6  &2.6 &74.9   &-                & 74.9& 35.4  &9.6      &67.8 &  -    &21.4 & -  &4.1  &15.5  &31.4&36.6&-\\

DPR~\cite{tsai2019domain}  
& 72.6   & 29.5 &77.2    & 3.5 & 0.4   &21.0 & 1.4  &7.9 &73.3    &-                & 79.0 & 45.7  &14.5      &69.4 &  -    &19.6 & -  &7.4  &16.5  &33.7 &39.6&-\\

PyCDA~\cite{lian2019constructing}       & 80.6   & 26.6  &74.5    & 2.0 & 0.1   &18.1 & 13.7  &14.2 &80.8    &-                & 71.0 & 48.0  &19.0      &72.3 &  -    &22.5 & -  &12.1  &18.1   &35.9 &-&-\\

CBST~\cite{zou2018unsupervised}      & 69.6   & 28.7  &69.5     & 12.1 & 0.1   &25.4 &  11.9  &13.6 &82.0   &-                & 81.9 & 49.1  &14.5      &66.0 &  -    &6.6 & -  &3.7  &32.4   &35.4&-&-\\

PIT~\cite{lv2020cross}      & 81.7  & 26.9  &78.4    & 6.3 & 0.2   &19.8 & 13.4  &17.4 &76.7   &-                & 74.1 & 47.5  &22.4      &76.0 &  -    &21.7 & -  &\textbf{19.6}  &27.7   &38.1 &-&- \\

%\hline
%&\multicolumn{21}{c}{Adversarial-training based methods}\\
FADA~\cite{wang2020classes}      & 80.4  & 35.9  &\textbf{80.9}    & 2.5 & 0.3   &\textbf{30.4} &7.9  &22.3 &81.8   &-                & \textbf{83.6} & 48.9  &16.8    &\textbf{77.7} &  -    &\textbf{31.1} & -  &13.5  &17.9   &39.5  &-&- \\

\hline
FDA~\cite{yang2020fda}       & 84.2  & 35.1  &78.0    & 6.1 & 0.4   &27.0 &8.5  &22.1 &77.2   &-                & 79.6 & 55.5  &19.9     &74.8 &  -    &24.9 & -  &14.3  &40.7   &40.5 &-&- \\
CD-AM~\cite{yang2021context}       & 73.0  & 31.1  &77.1   & 0.2 & 0.5  &27.0 &11.3  &27.4 &81.2  &-                & 81.0 & \textbf{59.0}  &25.6     &75.0 &  -    &26.3 & -  &10.1  &47.4  &40.8 &-&-  \\

LDR~\cite{yang2020label}      & 73.7  & 29.6  &77.6   & 1.0& 0.4   &26.0 &14.7  &26.6 &80.6   &-                & 81.8 & 57.2  &\textbf{24.5}    &76.1 &  -    &27.6 & -  &13.6  &46.6   &41.1 &-&-  \\

SA-I2I~\cite{musto2020semantically}       & 79.1  & 34.0  &78.3   & 0.3 & 0.6   &26.7 &15.9  &29.5 &81.0   &-                & 81.1& 55.5  &21.9     &77.2 &  -    &23.5 & -  &11.8  &47.5   &41.5 &-&-  \\
%SAC~\cite{araslanov2021self}  &CVPR2021     & 90.0  & 53.1 &86.2   & 33.8 & 32.7   &38.2 &  46.0 & 40.3 &84.2    &26.4& 88.4 & 65.8   &28.0   &85.6 &  40.6  & 52.9 & 17.3 &13.7 & 23.8  &\textbf{49.9}  \\

CLAN~\cite{luo2021category}       & 82.0  & 33.7  &79.8   & - & - &- &6.4  &8.9 &78.7  &-                & 82.5 & 49.1  &12.9    &75.9 &  -    &21.9 & -  &5.1  &13.3  &-  &42.3&-\\

\hline
Source Only    & 2.6 & 16.3  &33.8   & 0.3 & 0.0  &18.7 &2.4  &4.7&71.2  &-                & 33.0 & 41.3 &1.1    &40.7 &  -    &7.5 & -  &0.3  &2.1  &\textbf{17.2} &19.7&14.5 \\
Ours~(1shot)     & 86.0  & 32.7  &75.9  & 7.4 & 0.7  &28.0&  17.5 &30.3 &83.3  &15.2
                    & 70.8 & 55.2 &6.4   &71.6& 0.0  &7.7 & 0.7 &2.3  &19.2  &37.2 &43.0 &32.2 \\
Ours~(5shot)    & \textbf{89.3}  & \textbf{43.4}  &79.1   & \textbf{15.4} & \textbf{9.9}  &29.3& \textbf{24.3} &\textbf{34.8} &\textbf{83.5}  &\textbf{28.8}
                    & 79.8 & 56.5  &12.0   &75.3& 1.6  &13.6 & 0.2 &9.4  &\textbf{47.7}  &\textbf{43.9} &\textbf{49.9} &\textbf{38.6} \\

\hline

\end{tabular}
\vspace{-3mm}
\label{table_sota_synthia}
\end{table*}

\subsection{SYNTHIA to Cityscapes}
\label{sec_synthia2cityscapes}
In the above experiments, GTA5 was the source domain. To show that our method can be generalized to different datasets, we also conducted experiments on the  SYNTHIA-to-Cityscapes adaptation task. Following common practice, we adopted the SYNTHIA-RAND-CITYSCAPES dataset~\cite{ros2016synthia} which contains 9,400 densely annotated images for our study. Different from GTA5, SYNTHIA shares only 16 common labels with the 19 classes present in the Cityscapes dataset. In other words, the SYNTHIA-to-Cityscapes adaptation is essentially an OSDA problem. Nevertheless, most existing methods consider a standard domain adaptation problem on this dataset and report mIoU on the 16 common classes. A few methods considered only 13 classes, where the wall, fence and pole were also excluded. In this experiment, we consider both the OSDA and standard case. For convenience, we denoted these mIoU values computed based on different class numbers by mIoU$N$ (e.g., mIoU16).

\noindent \textbf{Overall results}: Similar to the experiments on GTA5, we first trained the source only baseline for 50 epochs and achieved 17.2\% mIoU16, 19.7\% mIoU13, and 14.5\% mIoU19. Then, we trained our FSDA using different number of support images for the same number of epochs.  With 1-shot support images, we achieved 37.2\% mIoU16, 43.0\% mIoU13 and 32.2\% mIoU19. With 5-shot support images, we achieved 43.9\% mIoU16, 49.9\%mIoU13, and 38.6\% mIoU19.

\begin{figure}[!ht]
\centering
\includegraphics[width=3.5in]{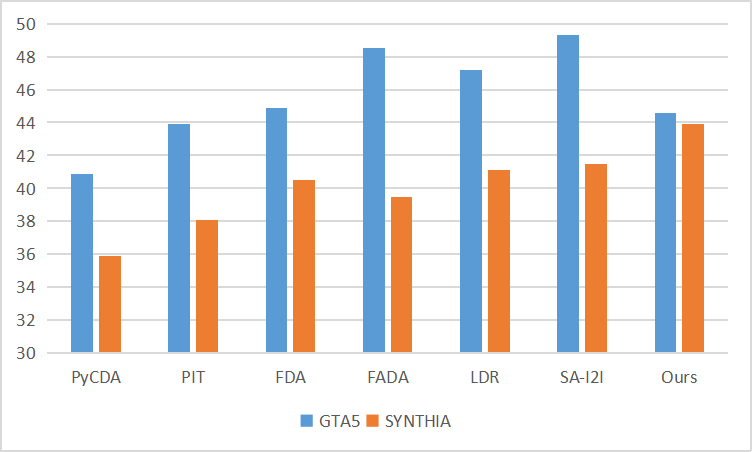}
\vspace{-4mm}
\caption{Performance of state-of-the-art methods on Cityscapes using different source domains. Most methods encounter a significant performance drop when the source domain changes from GTA5 to SYNTHIA. In contrast, our method has only a minor performance drop.}
\label{fig_GTA5TOSYTH}
\end{figure}

\begin{figure*}[!ht]
\centering
\includegraphics[width=7.1in]{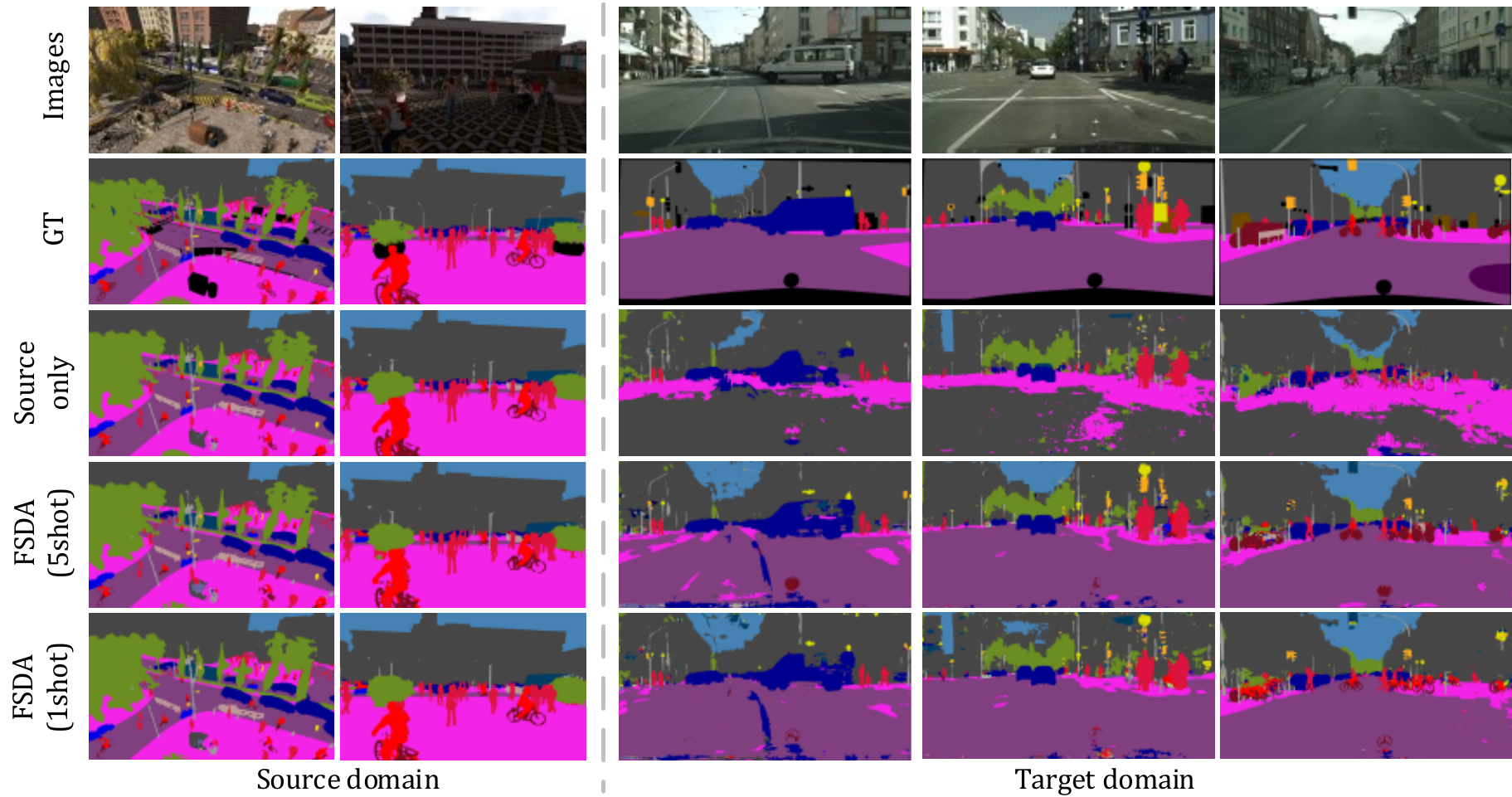}
\caption{Qualitative results on SYNTHIA-to-Cityscapes adaptation. The first two columns are the results on the source domain, and the last three columns are the results on the target domain.}
\label{fig_SYTHvisualization}
\end{figure*}

\noindent \textbf{Quantitative Analysis:} We first analyze our results from the perspective of standard domain adaptation, i.e., considering the mIoU16 for analysis. It can be observed from Table \ref{table_sota_synthia} that our FSDA achieved the state-of-the-art performance on SYNTHIA-to-Cityscapes adaptation task. Moreover, we see that the 17.2\% mIoU16 of source only baseline achieved with SYNTHIA lags far behind that obtained with GTA5 (i.e. 24.0\% mIoU16 derived from the 21.8\% mIoU19 in Table \ref{table_results_baseline}), even though the number of source images of SYNTHIA is significantly larger than that of GTA5 (9,400 vs 2,500).
This is mainly because many images of SYNTHIA were sampled from the same scene with different lighting or weather conditions, making the domain shift with respect to Cityscapes much larger than that of GTA5. As a result, the performance degradation is also visible in state-of-the-art domain adaptation methods when the source domain is changed  to SYNTHIA from GTA5 (see Fig. \ref{fig_GTA5TOSYTH}). In particular, the performance in terms of mIoU16 of PyCDA, PIT, FDA, FADA, LDR and SA-I2I dropped by 5.0, 5.8, 4.4, 9.0, 6.1 and 7.8 percentage points, respectively. In contrast, our method suffered only a 0.7 percentage point drop in performance. This shows that our method is more robust to the data distribution of source domain.

From the perspective of OSDA, our FSDA achieved 28.8\%, 1.6\% and 0.2\% mIoU on the three private classes of Cityscapes namely, terrain, truck and train. Although not very impressive, this is acceptable when compared to few-shot only baseline, which achieved 26.6\%, 3.8\% and 0.0\% mIoU on these three classes (see Table \ref{table_results_baseline}).
%\section{Discussion}

\noindent \textbf{Qualitative Analysis:} We also provide qualitative results for experiments on the SYNTHIA-to-Cityscapes adaptation task. As shown in Fig. \ref{fig_SYTHvisualization}, we applied the source only baseline and two variants of our method to SYNTHIA and Cityscapes dataset. Due to large distribution shifts, the source only baseline trained on SYNTHIA tends to segment large area of the road as building when adapted to the Cityscapes dataset. In contrast, our method can  distinguish the road and building in most cases, and perform much better on many other objects like pole, traffic sign, and traffic light.  In addition, the results on SYNTHIA images show that our method can achieve almost the same performance with the source only baseline, indicating that our method can reserve its representation ability on source domain after the domain adaptation, which is useful when it is required to applying the same model to both the source and target domains.

\section{Conclusion and Future Work}
In this paper, we introduced a simple generic framework for domain adaptation called FSDA. We showed that the FSDA can be generalized to many variants of domain adaptation tasks such as OSDA and MSDA with the support of a few annotated images from the target domain. In addition, our FSDA, developed from prototype-based few-shot learning, also provides a solution to few-shot learning on images that contain arbitrary number of objects.  The proposed FSDA achieved competitive performance (39.4\% mIoU) on the GTA5-to-Cityscapes adaptation task, and state-of-the-art performance (43.9\% mIoU16) on the SYNTHIA-to-Cityscapes adaptation task.
To the best of our knowledge, our method is the first to unify the domain adaptation and few-shot learning into a single framework.
%Thus, there are still many interesting topics worth exploring in the future.

Since the prototype-based method we used has been proven to be effective in few-shot learning, this paper mainly focused on domain adaptation task while paying less attention to apply the unified framework to few-shot segmentation. Nevertheless, current few-shot segmentation methods are still limited to images of low-resolution and containing only a few objects. Thus, an interesting direction is to extend our method to few-shot segmentation on images containing arbitrary number of classes. Many novel strategies (e.g., adaptive prototype learning~\cite{li2021adaptive}, self-guided learning~\cite{zhang2021self}) have been proposed to improve the performance of prototype-based few-shot segmentation in the past several years.  Hence, another promising future work is to validate whether such strategies can yield better domain-invariant prototypes when incorporated into our framework.

We have also tried the ResNet~\cite{he2016deep} as the backbone in our experiments. However, it did not yield better results compared with using VGG. We hypothesize this mainly caused by the batch normalization (BN)~\cite{ioffe2015batch} layer used in ResNet. The BN essentially computes the statistical properties of the features during the entire training process. Compared with existing methods that adopt all target images for training, our method access only a few target images in the training process. Thus, it is difficult for the BN layer to learn effective feature statistics of the target domain. To solve this problem, a future direction can be to perform unsupervised feature learning on target domain before applying our method.

\bibliography{mybibfile_nodoi}
%\end{thebibliography}

% \fi
\IEEEpeerreviewmaketitle

\begin{IEEEbiography}[{\includegraphics[width=1in,height=1.25in,clip,keepaspectratio]{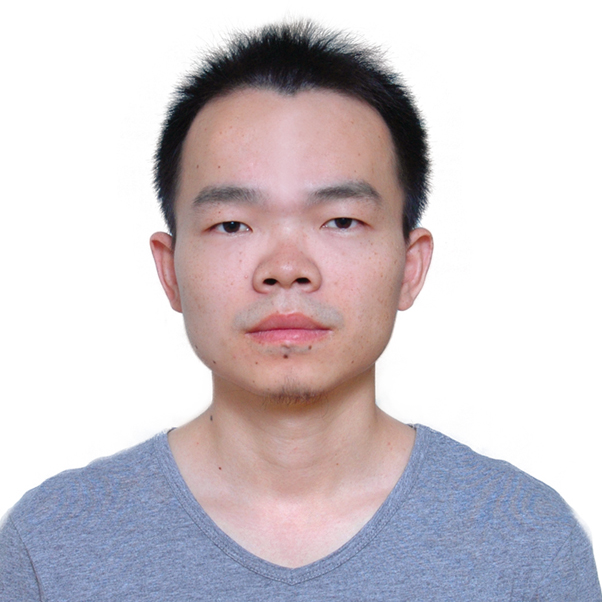}}]{Zhengeng Yang} received the B.S. and M.S. degrees from Central South University, Changsha, China, in 2009 and 2012, respectively. He received the Phd degree from Hunan University, Changsha, China, in 2020. He is currently a post-doctor researcher at Hunan University, Changsha. He was a Visiting Scholar with the University of Pittsburgh, Pittsburgh, PA during 2018 -2020. His research interests include computer vision, image analysis, and machine learning.

\end{IEEEbiography}

\begin{IEEEbiography}[{\includegraphics[width=1in,height=1.25in,clip,keepaspectratio]{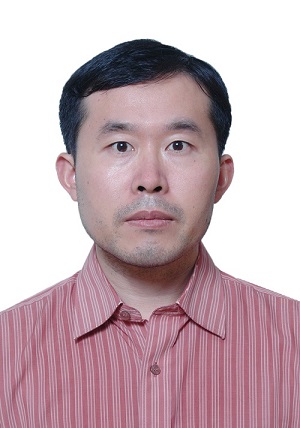}}]{Hongshan Yu} received the B.S., M.S. and Ph.D. degrees of Control Science and Technology from college of electrical and information engineering of Hunan University, Changsha, China, in 2001, 2004 and 2007 respectively. From 2011 to 2012, he worked as a postdoctoral researcher in Laboratory for Computational Neuroscience of University of Pittsburgh, USA. He joined the Hunan University as Assistant Professor in 2007 and became Professor in 2018. His research interests include autonomous robot and machine vision.
\end{IEEEbiography}

\begin{IEEEbiography}[{\includegraphics[width=1in,height=1.25in,clip,keepaspectratio]{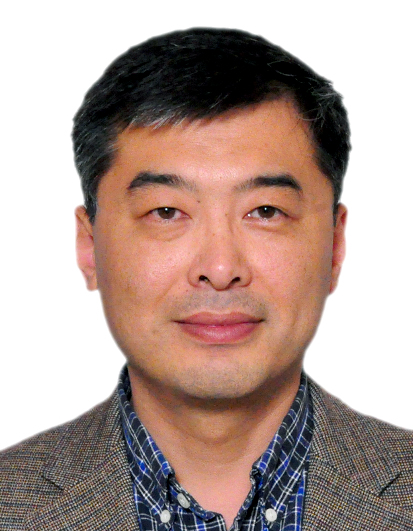}}]{Wei Sun} received the M.S. and Ph.D. degrees of Control Science and Technology from the Hunan University, Changsha, China, in 1999 and 2002, respectively. He is currently a Professor at Hunan University. His research interests include artificial intelligence, robot control, complex mechanical and electrical control systems, and automotive electronics.
\end{IEEEbiography}

\begin{IEEEbiography}[{\includegraphics[width=1in,height=1.25in,clip,keepaspectratio]{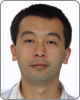}}]{Li Cheng} received the PhD degree in computer science from the University of Alberta, Canada.
He is a research scientist and group leader in Bioinformatics Institute (BII). Prior to joining BII July of 2010, he worked in Statistical Machine Learning group of NICTA, Australia, TTI-Chicago, and the University of Alberta, Canada. His research expertise is mainly on machine learning and computer vision. He is a senior member of the IEEE
\end{IEEEbiography}

\begin{IEEEbiography}[{\includegraphics[width=1in,height=1.25in,clip,keepaspectratio]{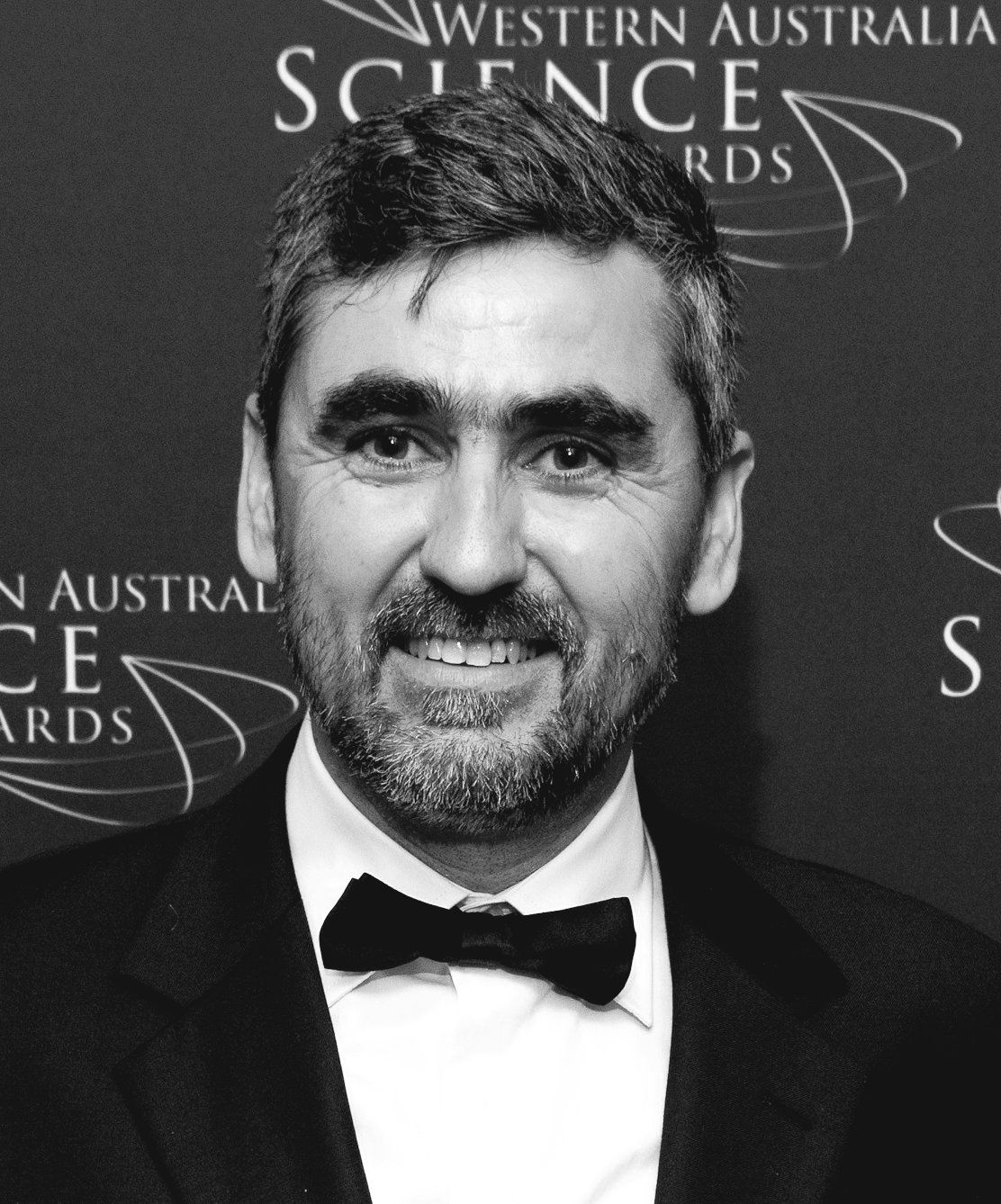}}]{Ajmal Mian} is currently a Professor of computer science with The University of Western Australia. His research interests include computer vision, machine learning, 3D shape analysis, human action recognition, video description, and hyperspectral image analysis. He has received three prestigious fellowships from the Australian Research Council (ARC) and several awards including the West Australian Early Career Scientist of the Year Award, the Aspire Professional Development Award, the Vice-Chancellors Mid-Career Research Award, the Outstanding Young Investigator Award, the IAPR Best Scientific Paper Award, the EH Thompson Award, and excellence in Research Supervision Award. He has received several major research grants from the ARC, the National Health and Medical Research Council of Australia and the US Department of Defence. He serves as an Associate Editor of the IEEE Transactions on Neural Networks and Learning Systems, the IEEE Transactions on Image Processing, and the Pattern Recognition Journal.
\end{IEEEbiography}

%
%% insert where needed to balance the two columns on the last page with
%%\newpage


%\vfill


\end{document}